%% file: main.tex
\documentclass[onecolumn]{tamplate/hexafuture}
\usepackage{tabularx,longtable}
\usepackage{fontawesome5}  

\usepackage{pgfplots}
\pgfplotsset{compat=1.18}
\definecolor{vlaLight}{HTML}{86B6EF}  
\definecolor{vaDark}{HTML}{256ABF}    
\definecolor{inkMuted}{HTML}{52514E}
\definecolor{evoVLA}{HTML}{86B6EF}   
\definecolor{evoBase}{HTML}{5598E7}  
\definecolor{evoGPT}{HTML}{256ABF}   
\definecolor{evoHex}{HTML}{104281}   
\newcommand{\vagap}[2]{
  \draw[black!40, line width=0.5pt] ([xshift=-14pt]axis cs:#1,101.5) -- ([xshift=-14pt]axis cs:#1,103.5)
        -- ([xshift=14pt]axis cs:#1,103.5) -- ([xshift=14pt]axis cs:#1,101.5);
  \node[font=\tiny, color=inkMuted, anchor=south] at (axis cs:#1,103.8) {$\Delta$\,\textminus#2};}

\title{Self-Evolving Coding Agents: From Digital Programs to Physical-World Intelligence}

\author{
\begin{center}
    Hongcheng Gao$^{*\dagger}$ \quad Jingjing Zhou$^{*}$ \quad Zelin Zheng \quad Shijia Ge \quad Jay Zhu \quad Yazhe Wang
    \\[5pt]
    Jianshu Zeng \quad Xuan Shangguan \quad
    Di Wu \quad Lingyu He \quad Zhiqi Jia 
    \\[5pt]
    Sihang Wu$^{\ddagger}$ \quad Xiao He$^{\ddagger}$ 
    \\[12pt]
    $^{*}$Equal Contribution \qquad $^{\dagger}$Project Lead \qquad $^{\ddagger}$Corresponding Author
    \\[10pt]
    {\sffamily\small\color{hexaprimary}\faGlobe\enspace\href{https://hexafuture.ai}{hexafuture.ai}}
    \vspace{-0.5cm}
\end{center}
}

\begin{document}
\abstract{%
Vision-language-action (VLA) and world-action (WAM) models map observations and instructions directly to robot actions. This directness ties a policy to training: minor layout or viewpoint changes cause failure, and instructions generalize poorly. The root cause lies in representation: task requirements, conditions, progress, and failure recovery are implicitly encoded in action sequences, making them difficult to inspect or revise. Digital coding agents offer a precedent: LLMs call tools, verify results, and revise from feedback as executable code. The same working pattern of explicit state, manageable execution, and revisable procedures underlies generalization and long-horizon execution in the physical world, letting physical experience return as reusable programs, memory, or evidence. We propose \textbf{Physical Coding}, representing task state and execution as code. \emph{Code as World} records objects, relations, constraints, and progress; \emph{Code as Policy} organizes planning, verification, recovery, and execution. We build HexaAnything, which calls perception, planning, and control tools, including VLA/WAM policies, and makes in-the-loop decisions from external feedback. Verified traces become data and memory, enabling evolution from tools and Harness to model weights, architectures, and ultimately hardware and task design. On RoboCasa365, HexaAnything improves Composite-Unseen and overall success over XR-1 VLA, and its Harness-trained HexaModel beats the base on every split, indicating code traces internalize physical execution. On PhyBench and a dual-arm AgileX robot, the agent autonomously completes physics experiments and most tabletop tasks, often faster than published results. We observe data, model, and tool self-evolution; future work targets weight internalization, autonomous redesign of architectures, languages, representations, and tasks, and deployment in manufacturing and science.
}

\maketitle
\justifying
\setlength{\parindent}{0pt} 
\section{Introduction}

Vision--language--action (VLA) and world--action (WAM) systems connect foundation models to physical interaction~\cite{zitkovich2023rt2,kim2024openvla,octo2024,black2024pi0,bjorck2025groot,li2026lingbotva}. They map visual observations and language instructions to robot actions, providing a compact interface between perception and control. How much of this mapping depends on the instruction is less clear. A QwenGR00T policy trained on LIBERO with the instruction masked succeeds in 92.3\% of trials, against 96.2\% for its instruction-conditioned counterpart (Section~\ref{sec:rw-vla}): when the scene determines the task, the policy maps scenes to trajectories. Released VLAs are similarly insensitive to the instruction, and they lose more than half of their success under small changes in viewpoint or initial robot state and approach zero when object layout is perturbed~\cite{fei2025liberoplus,zhou2025liberopro,li2024simpler}; increasing data or model capacity does not reliably remove this failure mode~\cite{li2024robovlms}. The limitation is structural rather than purely a matter of action capacity: a long-horizon task requires the system to track objects, constraints, unfinished subgoals, state changes, completion conditions, and recovery decisions, and none of these is represented in an action chunk and a stop signal. Closed-loop systems feed observations back to a planner~\cite{huang2022inner,yao2023react}, and structured embodied agents maintain state records or spatial constraints~\cite{yoneda2023statler,rana2023sayplan,huang2023voxposer,huang2024rekep}; yet these representations, tool semantics, and recovery rules are usually designed outside the learning loop, so feedback may improve one episode without becoming a reusable, independently verifiable artifact.

This diagnosis shifts the problem from ``can the model predict a better action?'' to ``does the system maintain a representation that can be inspected and revised throughout execution?'' The missing capability is therefore not another action primitive, but an interface that exposes task state, execution, and feedback as objects that can be checked and changed. Digital coding agents provide a useful precedent: large language models use code to represent procedures, call external tools, inspect intermediate results, and revise programs from feedback~\cite{chen2021codex,yang2024swe,wang2025openhands}. Code is compositional, executable, verifiable, versionable, and revisable, making it a natural interface between representation, execution, and feedback. Physical execution makes this interface more demanding because actions may be delayed, partially observed, noisy, or irreversible. A physical coding agent therefore needs a world representation that can be updated from images, depth, proprioception, and tool outcomes; a policy representation that can branch, loop, interrupt, and recover; and a verifier independent of the model's completion claim. The system must retain the observation that motivated an action, the state condition that authorized it, the tool outcome, and the evidence used to accept or reject it. The Harness connects these components and turns provenance into targeted revisions and reusable execution records.

\begin{figure}[!tp]
\centering
\includegraphics[width=0.98\linewidth]{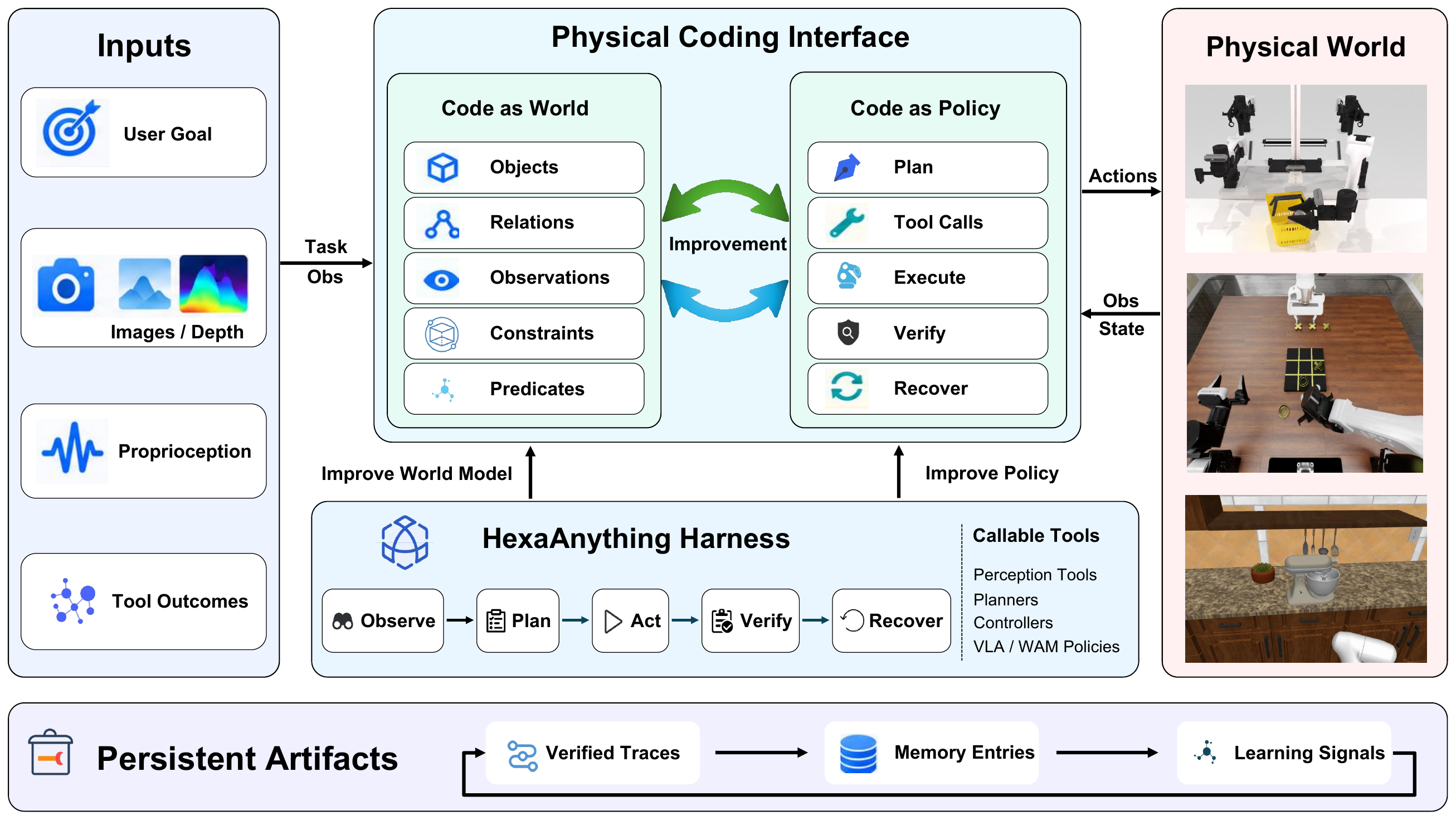}
\caption{\textbf{Physical Coding brings AI into the physical world through an explicit executable interface.} Code as World represents task-relevant objects, relations, observations, constraints, and progress predicates, while Code as Policy organizes planning, tool calls, action execution, verification, and recovery. Together they form the core interface through which the agent interacts with physical environments and accumulates reusable artifacts for continual improvement.}
\label{fig:bp-code-world}
\end{figure}

Based on this observation, we study \textbf{Coding Agents for the Physical World} and introduce \textbf{Physical Coding}: a formulation that represents both the relevant world state and the execution procedure as executable programs. \textbf{Code as World} describes objects, relations, observations, states, constraints, and progress predicates. \textbf{Code as Policy} organizes planning, tool calls, outcome verification, recovery, and action execution. Because these programs can be independently validated, edited, versioned, and rolled back, the experience produced by one physical execution need not disappear at the end of an episode: it can return to the system as a reusable program artifact, memory entry, or structured evidence for later updates. The system can therefore revise not only a policy checkpoint but also the world representation, the workflow, the tools and verifiers, the data used for learning, and eventually the model and the representation standards themselves (Section~\ref{sec:evolution-space}). Code as World exposes task-relevant state, constraints, and progress, while Code as Policy exposes the workflow connecting planning, action, verification, and recovery, as illustrated in Figure~\ref{fig:bp-code-world}.

This representation allows execution experience to produce persistent updates rather than only additional demonstrations. Successful executions can be validated and admitted as reusable Physical Coding records, memory entries, or learning signals for later system updates, while failed executions provide localized evidence for revising tools, workflows, predicates, and recovery procedures. These two feedback paths form the recursive \emph{model--Harness--environment} loop described in Section~\ref{sec:evolution-space}: interaction generates evidence, evidence updates the digital system, and the updated model and Harness determine the next round of physical interaction.

The resulting program has four coupled evolution targets---the Harness, the model, data and environments, and the embodiment and compute substrate (Section~\ref{sec:evolution-space})---connected by the typed execution trace, which determines what evidence is collected and where it can be reused. The program proceeds in three stages: Stage~1 bootstraps the coding--action--state--feedback loop; Stage~2 closes the data--model--Harness loop; and Stage~3 transfers it to constrained physical workflows with real sensors, users, engineers, and safety gates (Section~\ref{sec:roadmap}).

To instantiate this program, we develop \textbf{HexaAnything}, a physical coding agent that integrates state observation, task planning, action tools, verification, recovery, and feedback collection. Its action tools range from general-purpose robot tools to learned VLA/WAM policies, while the Harness maintains task state, coordinates execution, verifies intermediate outcomes, and adapts subsequent actions. We evaluate it on long-horizon robotic manipulation, simulated scientific experiments, and a real dual-arm robot. The experiments cover the Harness, tool revision, and a first data-to-model update: on manipulation, they test explicit workflow, verification, and recovery while holding the underlying action model fixed, and then train a new model on the traces the Harness returns. 

We provide preliminary evidence for physical self-evolution at the Harness, tool, and data-to-model levels. The experiments demonstrate controlled, partial improvements through verified execution traces, while full autonomous co-evolution of models, representations, environments, and embodiments remains future work.

Our contributions are summarized as follows:





\begin{enumerate}
    \item We formulate \textbf{Coding Agents for the Physical World} and introduce \textbf{Physical Coding}, an executable interface between physical state, action, evidence, and revision, instantiated through the coupled representations \textbf{Code as World} and \textbf{Code as Policy} that unify physical-world modeling and action execution.

    \item We develop \textbf{HexaAnything}, a physical coding agent that integrates language models, state observation, verifiers, and action tools into a unified physical execution loop.

    \item We provide evidence for Harness, data, model, and tool improvement on long-horizon manipulation, and show on the proposed \textbf{PhyBench} simulated laboratory benchmark that the same Harness autonomously completes scientific experiments, thereby motivating a staged path toward broader self-evolution.
\end{enumerate}

\section{Background: From Action Models to Coding Agents}

Our argument passes through five bodies of work: action models, code as policy, code as world, coding agents and their harnesses, and self-evolving agents. We review them in this order; each subsection ends with the point on which our formulation departs. An extended version with the mechanism-level comparison in Table~\ref{tab:related-work} is given in Appendix~\ref{app:related-work}.

\subsection{Action Models: VLA and WAM}
\label{sec:rw-vla}

\begin{figure}[t]
\centering
\begin{minipage}[b]{0.42\linewidth}
\centering
\begin{tikzpicture}
\begin{axis}[
    width=\linewidth, height=5.1cm,
    ybar=2pt, bar width=13pt,
    ymin=0, ymax=113, ytick={0,25,50,75,100},
    ylabel={Success rate (\%)}, ylabel style={font=\scriptsize, color=inkMuted},
    symbolic x coords={Spatial,Object,Long,Average},
    xtick=data, xticklabels={Spatial,Object,Long,\textbf{Average}},
    x tick label style={font=\scriptsize, yshift=-1pt},
    y tick label style={font=\scriptsize, color=inkMuted, /pgf/number format/assume math mode=true},
    enlarge x limits=0.15,
    axis x line*=bottom, axis y line=left, y axis line style={draw=none},
    x axis line style={black!55}, major tick length=0pt,
    ymajorgrids, major grid style={black!10},
    legend style={at={(0.5,1.02)}, anchor=south, legend columns=2, draw=none, font=\scriptsize, /tikz/every even column/.append style={column sep=8pt}},
    legend image code/.code={\fill[#1] (0cm,-0.08cm) rectangle (0.26cm,0.08cm);},
    nodes near coords={\pgfmathprintnumber[fixed,precision=1,fixed zerofill,assume math mode=true]{\pgfplotspointmeta}},
    every node near coord/.append style={font=\tiny, anchor=north, yshift=-1.5pt},
    clip=false,
]
\fill[hexasurface] ({rel axis cs:0.763,0}) rectangle ({rel axis cs:1,1});
\addplot[fill=vlaLight, draw=none, every node near coord/.append style={color=black!80}]
  coordinates {(Spatial,97.3) (Object,99.6) (Long,91.6) (Average,96.2)};
\addplot[fill=vaDark, draw=none, every node near coord/.append style={color=white}]
  coordinates {(Spatial,91.3) (Object,98.5) (Long,87.1) (Average,92.3)};
\legend{VLA, VA (no instruction)}
\vagap{Spatial}{6.0}\vagap{Object}{1.1}\vagap{Long}{4.5}\vagap{Average}{3.9}
\end{axis}
\end{tikzpicture}
\subcaption{Success rate on LIBERO.}
\label{fig:va-vla-bars}
\end{minipage}\hfill
\begin{minipage}[b]{0.555\linewidth}
\centering\scriptsize
\setlength{\tabcolsep}{0.7pt}\renewcommand{\arraystretch}{0.9}
\begin{tabular}{@{}cccc@{}}
\multicolumn{4}{@{}p{\linewidth}@{}}{\textbf{VLA}\quad prompt: ``pick up the alphabet soup and place it in the basket''\hfill\textcolor{green!55!black}{\ding{51}}} \\
\includegraphics[width=0.245\linewidth]{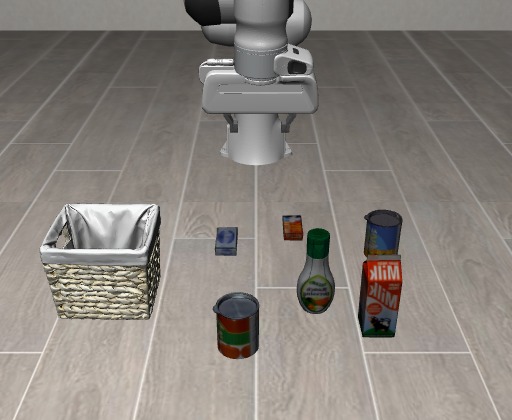} &
\includegraphics[width=0.245\linewidth]{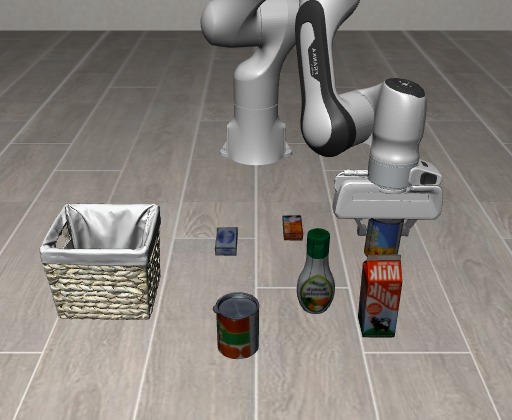} &
\includegraphics[width=0.245\linewidth]{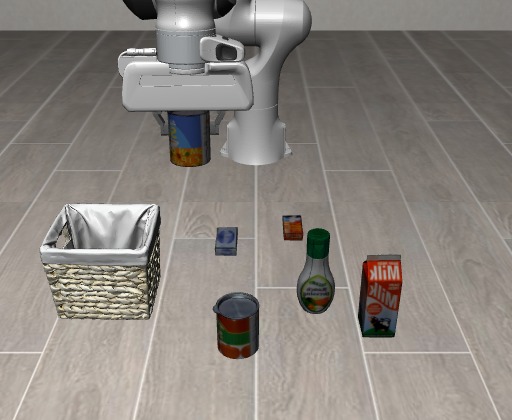} &
\includegraphics[width=0.245\linewidth]{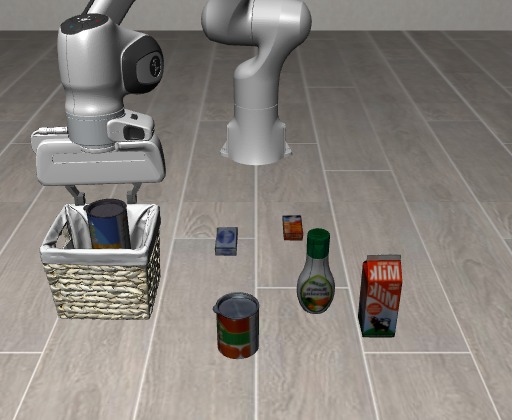} \\[3pt]
\multicolumn{4}{@{}p{\linewidth}@{}}{\textbf{VLA}\quad prompt: $\langle$empty$\rangle$ \hfill\textcolor{green!55!black}{\ding{51}}} \\
\includegraphics[width=0.245\linewidth]{figures/libero_rollout/obj_t0_start.jpg} &
\includegraphics[width=0.245\linewidth]{figures/libero_rollout/obj_t0_grasp.jpg} &
\includegraphics[width=0.245\linewidth]{figures/libero_rollout/obj_t0_carry.jpg} &
\includegraphics[width=0.245\linewidth]{figures/libero_rollout/obj_t0_place.jpg} \\
\color{inkMuted}start & \color{inkMuted}grasp & \color{inkMuted}carry & \color{inkMuted}place \\
\end{tabular}
\subcaption{Rollouts from the same LIBERO-Object initial state.}
\label{fig:va-vla-rollout}
\end{minipage}
    \caption{ \textbf{Removing the instruction barely changes behavior on LIBERO.} (a) Masking the instruction during training lowers success by at most 6.0 percentage points; both policies are QwenGR00T in StarVLA (Qwen3-VL-4B backbone, GR00T-style action head) trained jointly on the LIBERO training data, with 500 trials per suite. (b) Given the instruction or an empty prompt, the same VLA picks up the alphabet soup and places it in the basket. The VLA is a LIBERO-finetuned $\pi_{0.5}$~\cite{intelligence2025pi05}.}
\label{fig:va-vla}
\end{figure}

Vision--language--action (VLA) models map an image and an instruction to a chunk of robot actions~\cite{zitkovich2023rt2,kim2024openvla,black2024pi0,octo2024,bjorck2025groot}, yet much of this mapping does not depend on the instruction. A QwenGR00T policy (Qwen3-VL-4B backbone~\cite{bai2025qwen3vl}, GR00T-style action head~\cite{bjorck2025groot}) trained in StarVLA~\cite{starvla2026} on LIBERO~\cite{liu2023libero} with the instruction masked succeeds in 92.3\% of trials averaged over three suites, against 96.2\% for its instruction-conditioned counterpart, and trails it by at most 6.0 percentage points on any suite (Figure~\ref{fig:va-vla}); LangForce reports rates within 1.1 points of ours~\cite{lian2026langforce}. When the initial scene determines the task, the policy maps scenes to trajectories, and high success on such benchmarks does not show that it follows the instruction. Released VLAs are similarly insensitive to the instruction, yet lose more than half their success under small changes in viewpoint or initial robot state~\cite{fei2025liberoplus} and approach zero when object layout or task order is perturbed~\cite{zhou2025liberopro}.

World--action models (WAMs) add a video-prediction objective~\cite{yang2024unisim,li2026lingbotva}, which rewards plausible frames rather than correct physics: video models generalize by imitating the nearest training case and fail out of distribution~\cite{kang2024howfar}, and their visual realism does not track physical understanding~\cite{motamed2025physicsiq}. A predicted frame is also not a checkable state; whether it shows both ingredients inside the oven must still be judged by another model.

In both, decomposition, completion, and recovery are implicit in action chunks and a stop signal, so an oven closed with one ingredient outside leaves nothing in the interface to detect or repair. We keep the VLA or WAM as one action tool inside a program that owns these decisions (Section~\ref{sec:system-integration}).

\subsection{Code as Policy}
\label{sec:rw-code-as-policy}

Code as Policies made the program the policy: a language model writes Python that composes perception calls, control primitives, and control flow, and the program runs on the robot~\cite{liang2023code}, building on earlier work that grounded language plans in available skills~\cite{huang2022language,ahn2022can,singh2022progprompt}. CaP-X turns this line into a measurable framework~\cite{fu2026cap}. In CaP-Gym an agent writes programs over perception and control primitives; CaP-Bench evaluates twelve frontier models at several levels of abstraction; CaP-Agent0 adds multi-turn interaction, structured execution feedback, visual differencing, skill synthesis, and ensembled reasoning, and reaches human-level success on several tasks in simulation and on real robots. Its central finding is that success falls sharply once human-designed abstractions are removed, a dependence the authors call designer scaffolding.

Four properties of the program explain this dependence, each following from the previous one. First, the program's inputs are the outputs of perception primitives (boxes, masks, poses), and no node in the program checks them: perception is trusted, not verified. Second, errors therefore compound along the program. A long-horizon task is a chain of primitives whose failure probabilities multiply, and without a state predicate a drift in the middle of the chain surfaces only at the end. Third, improvement can act only on how primitives are composed, through skill synthesis, ensembling, and retries, so the accuracy of the primitives is a ceiling; visual differencing tries to add verification, but it asks a VLM to compare frames, which returns to the perceptual limits reviewed next. Fourth, the gains do not persist. Test-time strategies are paid for again in every episode, and CaP-RL updates weights from the gym's verifiable reward rather than from states verified during execution~\cite{fu2026cap}. Closed-loop variants such as Inner Monologue and ReAct feed observations back to the planner~\cite{huang2022inner,yao2023react}, and Voyager stores successful programs for reuse~\cite{wang2023voyager}, but in each the loop lives outside the program: verification is the model's reading of an observation, and recovery is a re-prompt.

The policy program $P$ in our formulation contains, beyond decomposition, tool selection, and action calls, verification, stopping conditions, and recovery as explicit nodes that can be checked statically. Verification is against the world program $W$ rather than against primitive outputs or the model's own judgment, and $P$ is edited from execution evidence and kept across episodes.

\subsection{Code as World}
\label{sec:rw-code-as-world}

Vision--language models answer semantic questions about images well and perceptual questions poorly. On counting, relative depth, and spatial relations, BLINK and Eyes Wide Shut report accuracy near chance and trace the failures to CLIP-style encoders that map visually distinct images to similar embeddings~\cite{fu2024blink,tong2024eyes}. Frontier models still fail simple geometric probes such as whether two lines intersect or how many circles overlap~\cite{rahmanzadehgervi2024blind}. Object hallucination follows language priors: models report objects that co-occur with the scene type rather than objects that are present~\cite{li2023pope}. Spatial understanding across viewpoints and over time is weaker still~\cite{yang2025thinking}, and closing these gaps has required dedicated spatial supervision~\cite{chen2024spatialvlm,song2025robospatial}. In robot settings the same errors appear as unreliable success detection and physical-property estimation~\cite{du2023success,gao2024physically}. A model asked whether both ingredients are inside the oven, or which plate is still missing a sausage, must count, localize, and judge containment, the operations on which these evaluations report the lowest accuracy. A textual answer from the VLM therefore cannot serve as evidence of state: it cannot be checked, and it is biased toward what the scene usually contains.

VCode uses the model differently. It casts image understanding as SVG generation: the model writes code whose rendering must preserve the symbolic content of the image, and fidelity is scored by whether a separate model can answer questions from the rendering~\cite{lin2025vcode}. Vector-graphics reasoning, image-to-SVG generation, and screenshot-to-code benchmarks use code the same way~\cite{wang2024vdlm,rodriguez2025starvector,si2024design2code}, and Visual Sketchpad lets a model draw with code during inference~\cite{hu2024sketchpad}. Three properties of this route matter for a world program. The representation is executable, so its errors are visible: a missing object, a wrong count, or a misplaced relation shows in the rendering or fails a predicate, whereas the same error in a textual answer is indistinguishable from a correct one. The representation is editable and accepts structured input: VCode's agent revises its SVG against rendering discrepancies and calls detectors and parsers for cues, which separates what the model reasons about from what a perception tool measures. And code is also the medium of the policy, so state and action share one representation that one verifier can check.

The case is stronger in embodied settings than in image benchmarks. A robot observes the same scene repeatedly, so a code representation can be updated incrementally, entry by entry, whereas a VLM answers each query from scratch and its answers can contradict one another without anyone noticing. Errors have physical consequences, so a representation that can be checked before an action is taken is worth more than one that is scored afterwards. And the representation must be built from what a real robot has: images, proprioception, and tool outputs. HexaAnything therefore writes the world program from images in the VCode manner and does not read simulator state. This contrasts with harnesses that verify against privileged simulator signals; Zetta, for example, reverts to internal simulator states, contact forces, and collision intensities when visual evidence is insufficient~\cite{ding2026zetta}, which is unavailable on a physical robot.

Structured state representations for embodied agents share parts of this design: VisProg and ViperGPT organize perception as programs~\cite{gupta2023visprog,suris2023vipergpt}, ConceptGraphs and SayPlan expose scene graphs to the planner~\cite{gu2024conceptgraphs,rana2023sayplan}, Statler keeps a state record updated after every action~\cite{yoneda2023statler}, VoxPoser and ReKep write spatial constraints as code over detected keypoints~\cite{huang2023voxposer,huang2024rekep}, and program-synthesized world models write the dynamics as code~\cite{tang2024worldcoder,bai2026visualpatchworld,chen2026codeworldmodel}. Each is built for one query or one episode. The world program $W$ in our formulation records task-relevant objects, relations, constraints, and progress predicates, each with the provenance of the observation behind it; it is checked by an independent verifier rather than by rendering fidelity; and it persists across episodes, so a failed execution can add a predicate or revise a constraint.

\subsection{Coding Agents and Harnesses}
\label{sec:rw-harness}

Software coding agents established that what an agent can do is set largely by the Harness around the model: the tools it exposes, the sandbox it runs in, and the tests it must pass~\cite{yang2024swe,wang2025openhands,chen2021codex}. Two elements of this setting carry over. Tests are a verifier independent of the model, which is why a software agent can be trusted to iterate; and the agent can edit its own scaffolding, as when Voyager grows a skill library~\cite{wang2023voyager} or autoresearch lets an agent modify the training script within a fixed evaluation protocol~\cite{karpathy2026autoresearch}. Neither element exists by default in a physical environment.

Embodied harnesses supply them around a frozen policy. Thea wraps robot capabilities as callable tools, keeps a scene graph as context, and reports action outcomes as exit codes~\cite{wang2026towards}. Guava runs a perception--reasoning--action loop over an LM and distills the resulting behavior into a 4B model from fewer than 2K simulated trajectories~\cite{liu2026guava}. Harness VLA wraps a frozen VLA as a retryable contact primitive, composes it with analytic primitives, and learns each primitive's operating range from execution traces~\cite{zhang2026harness}. Zetta runs three loops at different timescales, evolves critics and recovery skills online, and gates skill updates on validation rollouts~\cite{ding2026zetta}. SHAPER evolves a skill library and a context-code harness through target-environment rollouts~\cite{wang2026self}. GaP, BATON, and ASPIRE add graph-structured policies, transition-aware memory, and skill discovery under the same pattern~\cite{chen2026gap,xu2026baton,lu2026aspire}. These systems report large gains over the bare policy, and they show that critics, recovery rules, and skills can be revised without retraining.

The difference from our formulation lies in who learns. In each of these systems what is learned is stored outside the model, in memory, skill libraries, operating-range rules, or critics, and the model that reasons is held fixed; Zetta and SHAPER state this explicitly. Capability is then bounded by what retrieval and context can carry. In our formulation the Harness is also an instrument for collecting data: execution traces that pass an independent evaluator become Physical Coding data, and the model trained on them (HexaModel) becomes the planner of the next version. Two design choices follow from this. Because traces will be trained on, verification cannot come from the model itself: the verifier returns typed verdicts (pass, fail, insufficient evidence, blocked, safety stop) and every observation carries its provenance, whereas an exit code or a critic produced by the same model would confirm its own errors. And because edits will be inherited by the next version, what evolves is the typed workflow, its world predicates, and its tools, each admitted only after static checks, regression tests, and the official evaluator, rather than functions appended to a skill library.

\subsection{Self-Evolving Agents}
\label{sec:rw-self-evolving}

Prior work has made individual components of an agent adaptive. On the data side, GenSim, RoboGen, and CurricuLLM generate tasks and curricula~\cite{wang2024gensim,wang2023robogen,ryu2025curricullm}; MimicGen, GenSim2, RoboTwin, and HumanoidGen synthesize demonstrations~\cite{mandlekar2023mimicgen,hua2024gensim2,mu2025robotwin,jing2026humanoidgen}; Eureka, Text2Reward, DrEureka, and REvolve search reward code from rollouts~\cite{ma2024eureka,xie2024text2reward,ma2024dreureka,hazra2025revolve}; and RoboPlayground, AutoEval, and Eval-Actions automate evaluation~\cite{wang2026roboplayground,zhou2025autoeval,liu2026trustworthy}. On the model side, LoRA-style adapters update parameters cheaply~\cite{hu2021lora}, ENPIRE and Agent-Driven Autonomous RL let agent-written code steer training~\cite{xiao2026enpire,khandelwal2026agent}, and AutoML-Zero, AlphaEvolve, the Darwin G\"odel Machine, and The AI Scientist search over algorithms, programs, or experiments~\cite{real2020automl,alphaevolve,zhang2026darwin,lu2024ai}. On the substrate side, CompilerGym, MLGO, KernelBench, and CUDA Agent optimize compilers and kernels from execution feedback~\cite{cummins2022compilergym,trofin2021mlgo,ouyang2025kernelbench,dai2026cuda}, and Holodeck, SceneSmith, and SimFoundry generate simulator scenes~\cite{yang2024holodeck,pfaff2026scenesmith,ranawaka2026simfoundry}. Table~\ref{tab:related-work} groups these systems by the artifact they adapt, what they hold fixed, and the evidence they report.

Two observations organize this literature for our purposes. A training reward is not an independent verifier: when policy, reward generator, and scorer share data or are optimized jointly, gains can reflect reward exploitation rather than transferable capability~\cite{ma2024eureka,xie2024text2reward}. And a persistent artifact is not a model update: a skill stored in memory, a new tool route, or a revised Harness leaves the weights unchanged, and improvement after such a change does not show the model has learned~\cite{wang2023voyager,sygkounas2026memento,zhang2026playful}. Most systems therefore demonstrate adaptation of one component while the surrounding components, in particular the programming language, tool semantics, evaluator, and base weights, stay fixed.

We formulate self-evolution as a coupled system over task specifications $z$, environments $e$, world and policy programs $W$ and $P$, Harnesses $H$, model state $\theta$, verifiers $\phi$, data $D$, and memory $M$ (Section~\ref{sec:formulation}). The present report instantiates and evaluates the execution interface in $H$ together with a first update of $D$ and $\theta$, in which traces collected by the Harness train a new model (Section~\ref{sec:model-evolution}). The coupled view exposes four gaps in prior work: improvements are measured while most surrounding components are held fixed; gains in one component need not move the system-level bottleneck; improvements in simulation are hard to attribute once physical failures enter; and few systems sustain updates with independent validation, regression control, provenance, and rollback. These gaps motivate the verifier-centered development path that follows.

\section{Physical Coding}
\label{sec:formulation}

\begin{figure}[t]
    \centering
    \includegraphics[width=0.98\linewidth]{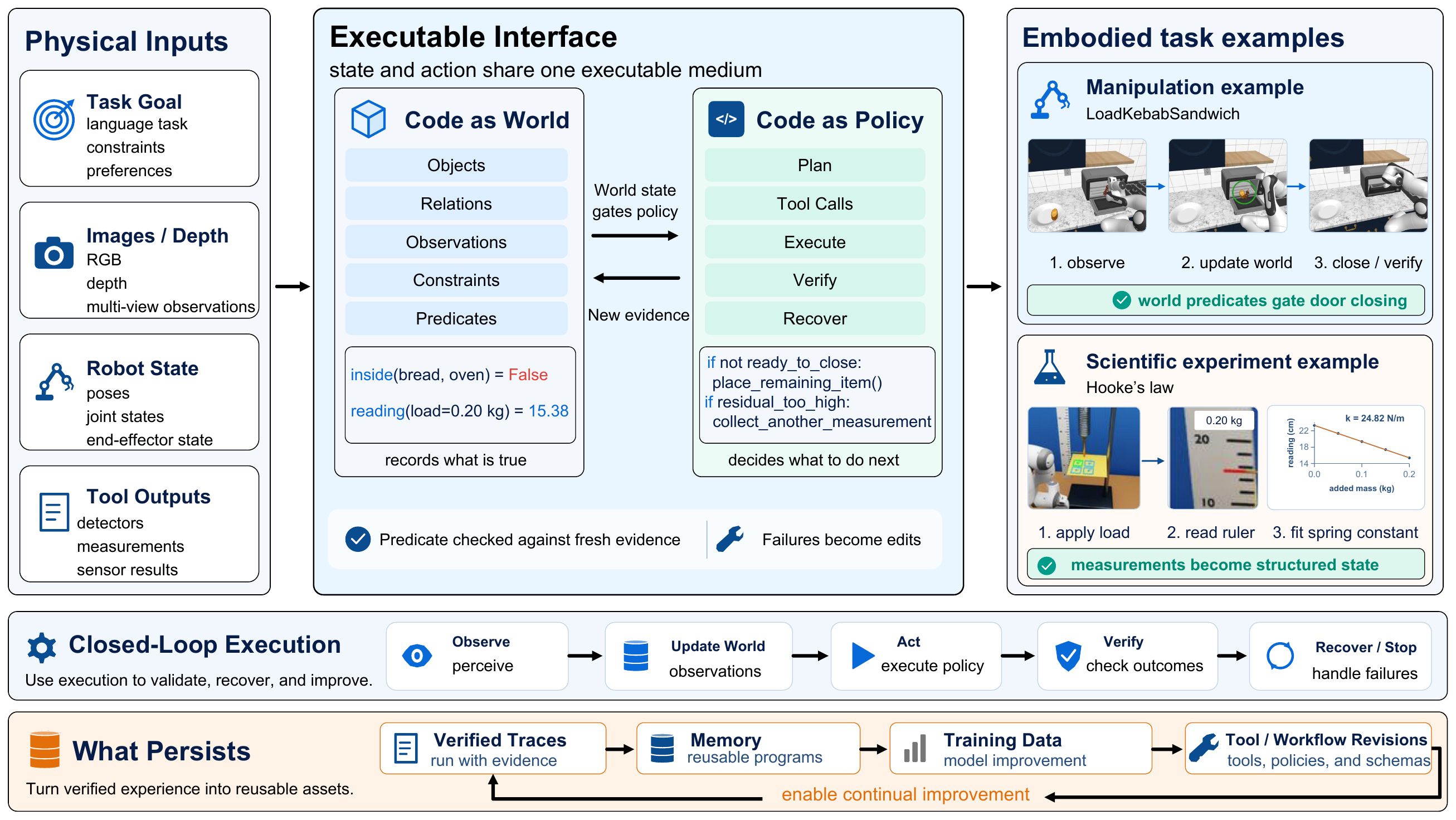}
    \caption{
\textbf{Physical Coding: coupling world state, policy execution, and evidence.}
Physical observations and tool outputs are translated into \emph{Code as World}, which represents task-relevant objects, relations, measurements, constraints, and progress predicates.
\emph{Code as Policy} uses this representation to organize planning, tool calls, action execution, verification, and recovery.
Execution produces new observations that update the world representation, while verification against fresh evidence determines whether to continue, re-observe, recover, or stop.
}
    \label{fig:Physical_Coding}
\end{figure}
\subsection{Why Physical Coding?}
\label{sec:why-physical-coding}

Coding agents have become capable in the digital world largely because of the medium they work in. A software task is already code: its state is the repository, an action is an edit or a tool call that returns a result, and completion is decided by tests that the agent does not write for itself~\cite{chen2021codex,yang2024swe,wang2025openhands}. The agent observes, acts, and checks within one executable medium, and what it learns persists in that medium: a fixed bug, a new tool, or a revised script is kept as a versioned edit, and trajectories that pass the tests can train the next model. This is what lets a coding agent improve through iteration rather than only through scale.

The physical world offers none of this by default. Its state is not a file that can be read, an action does not return a value saying whether it succeeded, and no test decides whether the task is done. Action models sidestep the problem by mapping pixels and instructions directly to motion,
\begin{equation}
    (\text{observations},\text{instruction}) \longrightarrow
    \text{action chunk} \longrightarrow \text{stop}
\end{equation}
so which subgoals remain, which conditions hold, whether the task is complete, and what to do after a failure are carried implicitly in the network and surface only as the next chunk or the stop signal. Section~\ref{sec:rw-vla} shows the consequence: a policy trained without the instruction nearly matches one trained with it, so much of what it learns is a fit from scenes to trajectories rather than an understanding of the task. Nothing in this interface can be inspected, corrected, or kept. A controller may finish moving while an object remains outside the target container, a model may stop while a required condition is false, and the only way to improve either is to collect more demonstrations. For a task of $n$ transitions with per-transition success probabilities $p_1,\ldots,p_n$, open-loop success is approximately $\prod_i p_i$; a system that checks the state after each transition can localize a failed one and, when the failure is recoverable, act on it rather than carry it forward.

Physical Coding gives the physical world the medium in which coding agents already work (Figure~\ref{fig:Physical_Coding}). The agent observes through code: it writes the task-relevant state as a program built from images, depth, sensor readings, and tool outputs, with predicates that can be evaluated. It acts through code: a program calls tools ranging from general-purpose robot tools to learned policies, with branches, interrupts, and recovery. It receives feedback through code: a verifier evaluates the program's predicates against evidence instead of accepting the model's claim that the task is done. Observation, action, and feedback then share one executable medium, as they do for a software agent, and so does improvement: a failed episode becomes a localized edit to a predicate, a tool, or a workflow, and a verified trace becomes data for the next model. Earlier code-based robot systems cover parts of this, mainly the policy side, with perception trusted and what is learned stored outside the model (Sections~\ref{sec:rw-code-as-policy} and~\ref{sec:rw-harness}). Nor is the loop specific to manipulation: a scientific experiment requires the agent to decide what to measure, operate instruments, and check whether the data support a conclusion (Section~\ref{sec:scientific-experiments}). \emph{Physical Coding} is thus the use of code for the state, the procedure, and the feedback of a physical task, so that the agent can improve through its own execution.

\subsection{Code as World and Code as Policy}
\label{sec:code-world-policy}

We use ``coding'' broadly to mean any behavior, rule, workflow, controller, evaluator, or environment transformation that can be expressed as an executable specification. A coding agent for the physical world needs two coupled representations. \emph{Code as World} is an executable account of the task-relevant world: its objects and relations, state variables, affordances, constraints, observations, and progress predicates. \emph{Code as Policy} is an executable account of how the agent acts in that world: how it decomposes a task, selects tools, calls actions, checks progress, recovers from failure, and replans. A robot motion primitive, a scene predicate, a reward function, a test suite, and a deployment workflow can all be expressed, inspected, and corrected in this form.

Neither half is sufficient on its own. Code as Policy without Code as World can call a controller but cannot reliably express whether an ingredient is still outside the oven or which plate is missing a sausage. Code as World without Code as Policy gives a structured description without a mechanism for acting, interrupting, or recovering. Together, they form an executable interface between state and action: intermediate progress becomes inspectable, failures become localizable, and proposed modifications become testable. In this sense, coding is a substrate for physical-world intelligence, not merely a convenient output format for a language model.

Two tasks from the evaluation make the pair concrete (Figure~\ref{fig:Physical_Coding}, right). In LoadKebabSandwich, where a kebab and a bread must go into an oven before its door is closed, the world program records which items are inside the oven and states ``both ingredients are inside'' as a predicate; the policy program places the items and allows the door to close only when that predicate holds (Section~\ref{sec:case}). In the Hooke's-law experiment of PhyBench, the world program records the apparatus, the weights on the tray, and each ruler reading together with the load that produced it; the policy program chooses which loads to apply, operates the arm to apply them, reads the ruler, and fits the stiffness from the recorded pairs (Section~\ref{sec:scientific-experiments}). In both, the task is stated over the world program and carried out by the policy program.

\paragraph{Core principle.} A physical-world coding agent maintains a pair of executable artifacts: a world program that records what is true and a policy program that decides what to do next. An independent verifier checks whether either program, or an edit to either program, is supported by evidence. The point is not simply to generate a policy in code, but to make the connection between physical state, action, and evidence explicit.

\subsection{How the Physical World Becomes Code}
\label{sec:how-physical-coding}

Physical Coding rests on two translations: from sensing to the world program, and from physical capability to the policy program. Feedback then closes the loop between them, and what is learned persists as code.

\paragraph{From sensing to Code as World.} World entries come from three sources. Scene structure---objects, regions, and relations---is written as code from images, in the manner of VCode (Section~\ref{sec:rw-code-as-world}); the agent does not read simulator state. Measurements come from tools: perception tools return detections, depth, and poses, and instruments return readings, such as the ruler, optical timer, and displacement recorder of PhyBench, with images cropped or enlarged on demand when a reading is hard to see. Execution state comes from the robot and the Harness: proprioception, gripper state, and the return value of every tool call. Each entry carries its provenance: the call that produced it, when it was observed, and, for a measurement, the experimental condition under which it was taken. Because the robot observes the same scene repeatedly, entries are updated one at a time rather than regenerated, so a new observation that contradicts an old entry is visible instead of silently replacing it. Task requirements are then written as predicates over these entries: both ingredients are inside the oven; each plate holds one bread and one sausage; a pendulum period is timed over complete cycles, in free motion, at an amplitude below $5^\circ$. When the available entries do not determine a predicate, the result is insufficient evidence rather than a guess, and the policy must observe again.

\paragraph{From physical capability to Code as Policy.} What the robot can do is exposed as typed tools: perception, motion planning, grasping, placement, and contact control, together with terminal, file, and Python tools for computation (Section~\ref{sec:scientific-experiments}), and learned action models called with a local instruction (Section~\ref{sec:harness-effect}). On the real robot the same tools are exposed through URAI and shared by humans and agents (Section~\ref{sec:real-world}). A policy is a program over these tools. Its nodes observe, act, verify, branch, loop, and recover; its conditions are predicates of the world program; and because the node kinds are typed, a proposed workflow can be checked before it runs (Section~\ref{sec:execution-contracts}). The program also holds what an action sequence cannot: a plan that fixes which experimental conditions to create and how they will be analyzed, the analysis code that turns readings into an estimate, and the recovery branch for a dropped item or a stalled call.

\paragraph{Closing the loop.} After each tool call the agent observes again, updates the world program, and evaluates the predicates the call was meant to change. The verdict, not the tool's own status or the model's claim, decides whether to continue, observe again, recover, or stop; the episode succeeds only when the goal predicates hold on fresh evidence. Section~\ref{sec:hexanything-harness} gives the runtime definition of this loop.

\paragraph{What persists.} Every step leaves a record in code: the world entries it read, the policy node that ran, the tool outcome, and the verdict. A failure can therefore be traced to a predicate, a condition, a tool, or a recovery branch, and fixed by editing that artifact and testing the edit against its parent (Section~\ref{sec:evolution-space}). Tools themselves are code and can be revised against a fixed evaluator (Section~\ref{sec:tool-evolution}); on the real robot, a programming agent writes, validates, and freezes them between episodes. Verified traces become training data for the next model (Section~\ref{sec:model-evolution}). This is how physical execution, once written as code, becomes material for self-evolution.

\subsection{Physical Coding Agents: Formulation and Scope}
\label{sec:physical-coding-agents}

We use \emph{Coding Agent for the Physical World} to denote the complete agent--environment interface, not merely a language model that emits robot programs. Such an agent maintains an executable account of the current physical state, selects and composes tools, observes the consequences of its actions, and decides whether to continue, recover, or stop. The model supplies general-purpose interpretation and synthesis; the Harness supplies typed tools, permissions, execution state, independent verification, and rollback. This separation is essential because physical completion is an external fact, not a statement generated by the same model that proposed the action.

We write $z$ for a task specification, $e$ for an environment, $W$ and $P$ for the world and policy programs, $H$ for the Harness that executes them, $\theta$ for the model state, and $V_\phi$ for a verifier; the runtime definition of an episode is given in Section~\ref{sec:hexanything-harness}. Self-evolution is the update of any subset of $(z,e,W,P,H,\allowbreak\theta,\phi,D,M)$, where $D$ is data and $M$ is memory, using the evidence from prior episodes. Edits to $W$ add predicates, observation abstractions, task constraints, or environment interfaces; edits to $P$ add workflows, tool compositions, recovery rules, or resource allocations; updates to $\theta$ internalize what these programs have supplied. The objective is not simply to maximize a scalar score, but to improve future performance while preserving safety, provenance, and reversibility. The verifier prevents a self-confirming story: a change is kept only if it improves held-out outcomes under an independently specified chain of evidence.

Our terminology has three levels. \emph{Physical Coding} is the representation and execution paradigm. \emph{Code as World} and \emph{Code as Policy} are its state and procedure components. \textbf{HexaAnything} is the physical coding agent studied in this report; its Harness implements the world--policy interface described here. The longer-term product direction is to expose this physical coding agent to engineers and operators in manufacturing, robotics, and scientific workflows. The experiments in this report cover the Harness, tool revision, and a first data-to-model update; coordinated evolution of all four targets in Section~\ref{sec:evolution-space} remains a longer-term program.

The remainder of the report follows this formulation. Section~\ref{sec:evolution-space} organizes what can evolve and the loop through which an update is admitted. Section~\ref{sec:hexanything-harness} describes HexaAnything and the Harness through which it realizes the world--policy interface. Sections~\ref{sec:preliminary-evaluation} and~\ref{sec:scientific-experiments} evaluate it on long-horizon manipulation and simulated scientific experiments, and Section~\ref{sec:real-world} reports its transfer to a real robot.

\section{Physical Self-Evolution}
\label{sec:evolution-space}
\begin{figure}[!tp]
    \centering
    \includegraphics[width=0.98\linewidth]{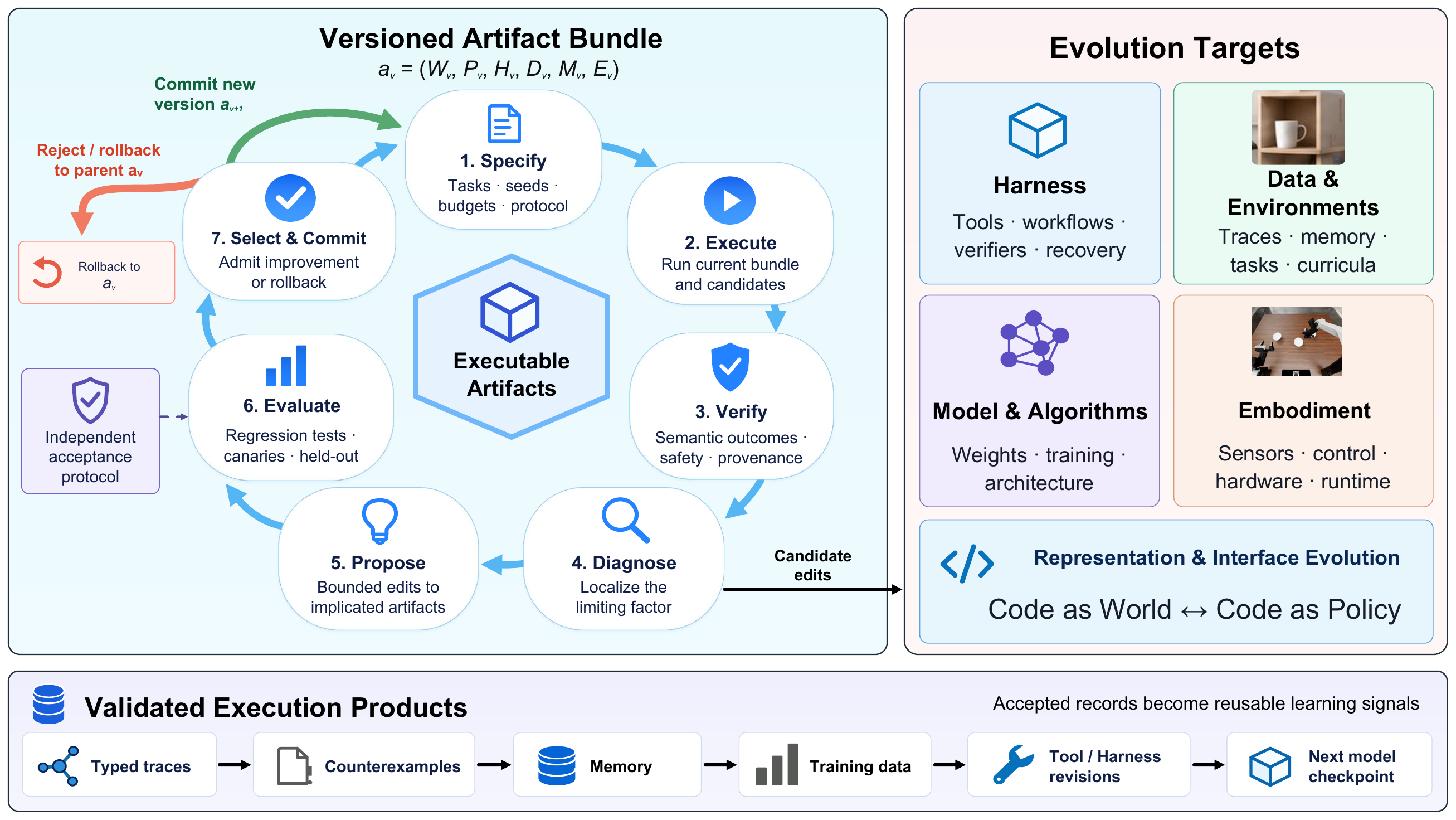}
    \caption{\textbf{Physical self-evolution.} Code as World and Code as Policy provide a shared, evolvable interface across four targets: the Harness, data and environments, model and algorithms, and embodiment and compute. Execution evidence guides candidate edits, independent evaluation determines acceptance or rollback, and validated records support subsequent learning and refinement.}
    \label{fig:Physical_Self_Evolution}
\end{figure}

Before describing our framework, it is useful to ask what current self-evolving systems actually change. The literature contains several successful but mostly component-local forms of evolution: environments, task generators, and curricula~\cite{wang2024gensim,wang2023robogen,nasiriany2024robocasa,ryu2025curricullm,xia2026sage}; demonstration and trajectory distributions~\cite{mandlekar2023mimicgen,hua2024gensim2,mu2025robotwin,jing2026humanoidgen}; rewards and training objectives~\cite{ma2024eureka,xie2024text2reward,ma2024dreureka,hazra2025revolve}; executable task programs and tool calls~\cite{liang2023code,mu2024robocodex,chen2024roboscript,li2024robocoder}; critics, skills, operating ranges, and recovery rules around a largely fixed base model~\cite{ding2026zetta,wang2026self,zhang2026harness}; model parameters and training procedures~\cite{hu2021lora,xiao2026enpire,khandelwal2026agent}; and, in a smaller set of systems, algorithms, programs, or robot morphology~\cite{real2020automl,alphaevolve,zhang2026darwin,bhatia2021evolution}. These results establish that nearly every component of an agentic stack can be made adaptive. They also reveal a common boundary: most systems evolve one category while treating the other categories as fixed interfaces. A task generator does not usually rewrite the policy representation; a policy optimizer does not usually change the evaluator or the tool schema; a hardware search is rarely coupled to the model and Harness that will operate on the resulting embodiment. The missing object is therefore not another isolated adaptive component, but an executable interface through which the whole stack can be inspected, revised, and re-evaluated.

Physical Coding is our proposal for that interface. Because Code as World and Code as Policy (Section~\ref{sec:code-world-policy}) are executable artifacts, a failure can be attributed to a missing state variable, an incorrect constraint, an unsuitable tool, an invalid workflow branch, or a weak verifier, and the corresponding artifact can be edited and tested; the world representation and the procedure acting on it become part of the search space, rather than code being merely a convenient policy output. Self-evolution in the physical world is therefore organized into four targets---the Harness, the model, data and environments, and the embodiment and compute substrate---with representation and interface evolution cutting across all four, and with evaluation and verification determining whether any update is admitted. This taxonomy also fixes the scope of the present report: its experiments cover Harness and tool evolution and a first data-to-model update, while coordinated evolution of all four targets remains a longer-term program.

\subsection{Overview}
The four targets are coupled by the executable world--policy interface rather than arranged as a one-way hierarchy. The Harness determines what the agent can observe, which tools it can call, how actions are verified, and how failures are recorded. Data and environments determine the experiences and counterexamples available to the system. The model uses admitted traces to internalize planning, tool use, state tracking, and recovery. The embodiment determines the sensors, actuators, timing, and physical constraints under which the other three targets must operate. An update to any one target can therefore change the evidence available to the others: a new verifier changes the training labels, a new model exposes different failure modes, a changed task distribution reveals gaps in the world representation, and a new embodiment changes the observation and action contracts.

The purpose of this organization is attribution as well as capability. A higher success rate is not by itself evidence that the model improved: it may result from a better recovery workflow, an easier task distribution, a more permissive evaluator, or a changed robot interface. Each candidate update must therefore identify its target, preserve the other relevant interfaces, and be compared against a versioned baseline under an independent evaluation protocol.

\subsection{The Self-Evolution Loop}
The common feedback mechanism follows the \emph{generate--evaluate--select} pattern shared by evolutionary program search and self-evolving coding agents: candidate artifacts are proposed from execution evidence, tested under a fixed protocol, and retained only when measured outcomes justify the change~\cite{real2020automl,alphaevolve,zhang2026darwin,lu2024ai,tao2024survey}. Figure~\ref{fig:Physical_Self_Evolution} organizes each round as an iterative cycle of \emph{experience acquisition, refinement, updating, and evaluation}~\cite{tao2024survey}: execution and diagnosis supply experience, proposed edits update the artifact bundle, and an independent evaluator gates admission.
The unit of evolution is a versioned artifact bundle
\(a_v=(W_v,P_v,H_v,D_v,M_v,E_v)\), containing the world representation, policy, Harness, admitted data, memory, and evaluator configuration. A candidate is not adopted on the strength of a single successful rollout; it is admitted when independent evidence shows that it improves the intended target without violating the contracts of the other components.

Each round has seven explicit operations (Figure~\ref{fig:Physical_Self_Evolution}):
\begin{enumerate}
    \item \textbf{Specify.} Freeze the task distribution, environment version, random seeds, budgets, and evaluation protocol for the round.
    \item \textbf{Execute.} Run the current artifact bundle and candidate variants, recording observations, state updates, tool calls, actions, verifier outputs, and resource use in a typed trace.
    \item \textbf{Verify.} Apply an evaluator independent of the proposing model to check semantic task predicates, safety conditions, and evidence provenance.
    \item \textbf{Diagnose.} Localize the limiting factor using trace inspection, counterexamples, ablations, and—where possible—matched comparisons. The diagnosis may point to a world representation, task specification, tool, workflow, verifier, model behavior, data distribution, or embodiment.
    \item \textbf{Propose.} Generate a bounded edit to the implicated artifact, together with its intended mechanism, affected interfaces, and predicted failure modes.
    \item \textbf{Evaluate.} Run static and type checks, regression tests, canary rollouts, and held-out evaluation. Candidate variants are compared with a frozen parent under the same seeds and budgets; evaluator changes are tested on an independent suite to prevent score inflation.
    \item \textbf{Select and commit.} Accept a candidate only if it satisfies the improvement, safety, provenance, and compatibility gates. Otherwise reject those that do not, keeping their traces as counterexamples. Improvements are retained, whereas rejected changes revert to the parent version. Accepted traces may be admitted to the Physical Coding corpus or used to update memory and the next model checkpoint.
\end{enumerate}

Formally, for version $v$ and task--environment pair $(z,e)$, the loop produces
\begin{equation}
  \xi_v \sim \operatorname{Execute}(P_v;W_v,H_v,e,z), \qquad
  o_v = V_v(\xi_v,z,e), \qquad
  a_{v+1} = \operatorname{Select}\!\left(\operatorname{Evaluate}(\operatorname{Propose}(a_v,o_v)),\mathcal{T}_{\mathrm{holdout}}\right),
\end{equation}
where $V_v$ is an independent verifier and $\operatorname{Select}$ requires improvement on a fixed hold-out suite, no regression on protected suites, valid provenance, interface compatibility, and a rollback-safe implementation. The loop can return different artifacts depending on the diagnosis: a failed placement may produce a recovery or tool edit; an incorrect completion claim may produce a stronger world representation or verifier; a recurring failure across tasks may produce a data or model update; and a failure tied to sensing or control limits may motivate an embodiment change. This evaluation gate is what allows the four evolution targets to improve without collapsing them into a single untraceable score, and it separates a better execution scaffold from a genuinely improved model or representation.

\subsection{Harness Evolution: Tools, Workflows, and Verification}
The Harness is the executable boundary between a coding agent and the physical environment. It includes tool schemas, observation abstractions, Code-as-World predicates, Code-as-Policy workflows, action routing, stopping conditions, recovery procedures, verifiers, memory interfaces, and resource allocation. A failure can therefore motivate a new perception or control tool, a different tool composition, a stronger state predicate, a revised termination rule, or a safer recovery branch. Existing embodied harnesses already evolve critics, skills, operating ranges, and recovery rules without retraining the base model~\cite{ding2026zetta,wang2026self,zhang2026harness}; software agents similarly demonstrate that tools, tests, and scaffolding strongly determine capability~\cite{chen2021codex,yang2024swe,wang2025openhands}. Their common limitation is that the reasoning model and much of the Harness are held fixed. In our formulation, verified failures are structured as candidate edits to both Code as World and Code as Policy, admitted only after static checks, regression tests, independent evaluation, and rollback.

In particular, Code as World is not a fixed ontology. The agent may eventually discover new object types, relations, affordances, observation abstractions, progress predicates, and task constraints when the existing representation cannot explain a failure. Code as Policy can likewise acquire new control decompositions, tool compositions, branching conditions, verification schedules, and recovery programs. This is the distinctive evolutionary opportunity created by making the world--policy interface executable: the system can improve not only the policy that acts in a representation, but also the representation that determines what can be observed, verified, and acted upon.

\subsection{Data and Environment Evolution}
Data and environment are themselves objects of evolution, not merely passive infrastructure. The data side includes demonstrations, trajectories, observations, memories, counterexamples, task distributions, and training corpora; the environment side includes task generators, curricula, simulators, reset conditions, reward functions, domain-randomization schemes, and evaluation suites. Systems such as MimicGen, GenSim2, RoboTwin, and HumanoidGen change demonstrations or trajectory distributions~\cite{mandlekar2023mimicgen,hua2024gensim2,mu2025robotwin,jing2026humanoidgen}; GenSim, RoboGen, RoboCasa, CurricuLLM, and SAGE change tasks, scenes, curricula, or simulators~\cite{wang2024gensim,wang2023robogen,nasiriany2024robocasa,ryu2025curricullm,xia2026sage}; and Eureka, Text2Reward, DrEureka, and REvolve change rewards or training objectives~\cite{ma2024eureka,xie2024text2reward,ma2024dreureka,hazra2025revolve}. Physical Coding connects these objects to the execution loop through a common typed trace. Simulation supplies inexpensive resets, controlled perturbations, counterexamples, and local evaluators, whereas physical environments provide sensor noise, calibration error, contact variation, latency, and failures that simulation may omit. Verified traces can update memory, refine task and objective generators, train a new checkpoint, or expose a missing tool or world representation.

\subsection{Model and Algorithm Evolution}
The model target contains the learned behavior that can eventually internalize capabilities first supplied by the Harness. It spans inference policies and prompts, tool-use and planning behavior, parameter-efficient adapters and post-training, full parameter updates, training objectives, and—at a longer timescale—architecture, routing, memory, action heads, and context allocation. LoRA provides a practical parameter-update mechanism~\cite{hu2021lora}; ENPIRE and agent-driven training place parts of the optimization loop under agent control~\cite{xiao2026enpire,khandelwal2026agent}; AutoML-Zero, AlphaEvolve, the Darwin G\"odel Machine, and The AI Scientist show that algorithms, programs, and research procedures can themselves become search objects~\cite{real2020automl,alphaevolve,zhang2026darwin,lu2024ai}. What these efforts do not yet establish for physical coding agents is sustained, independently evaluated co-evolution of workflow, world representation, model weights, and architecture. Our model-update claim is consequently staged: verified traces first train a new checkpoint, then reduced Harness assistance and held-out compositions test whether the capability has been internalized.

\subsection{Representation and Interface Evolution}
The executable interface itself can also evolve. At the task level, the system may refine the ontology, state variables, affordances, constraints, progress predicates, and abstractions used by Code as World. At the procedure level, it may extend the Physical Coding language: new syntax for branching or temporal conditions, new typed primitives, tool and verifier schemas, effect annotations, resource contracts, or compiler and runtime support. The relevant object is not natural language as such, but a representation standard that the model discovers to be useful for a task and an embodiment. It may be a symbolic schema, typed temporal record, executable DSL, learned latent code, state--action algebra, or protocol for aligning observations with world-program entities and physical effects. Most current systems keep these standards and modality boundaries fixed; Physical Coding makes them versioned executable interfaces that must compile or instantiate, preserve provenance, pass regression and safety tests, and improve held-out physical outcomes before adoption.

\subsection{Embodiment and Hardware Evolution}
The fourth target is the physical and computational substrate on which the Harness and model operate. It includes sensors, actuators, robot morphology, calibration, controller interfaces, training and inference accelerators, GPU memory, interconnects, storage, networking, and compute scheduling, as well as the hardware embodiment itself. Evolution Gym provides an example of searching over robot morphology~\cite{bhatia2021evolution}; compiler and kernel systems provide related examples of optimizing execution substrates from feedback~\cite{cummins2022compilergym,trofin2021mlgo,ouyang2025kernelbench,dai2026cuda}. In Physical Coding, both robot and compute changes are admitted through versioned interfaces: a modified embodiment must expose a compatible observation/action contract, while a modified compute substrate must preserve numerical behavior, training reproducibility, latency, energy, and cost constraints. Candidate changes must pass simulation, software, and physical safety tests and be evaluated with the model and Harness that will actually use them. Compute and hardware evolution are therefore longer-term targets, not assumptions of the current Harness-bootstrap experiments.

In every category, a successful episode enters the corpus only after independent verification, provenance checks, and contamination filtering; a failure becomes a localized counterexample for a representation, tool, workflow, verifier, model update, task generator, or embodiment. To separate these effects, we use an evidence ladder: E0 means that an artifact runs; E1 that an oracle can complete the task; E2 that it discriminates among policies; E3 that training on it improves a new checkpoint; and E4 that the resulting system transfers to a real robot or documented distribution shift~\cite{zhou2025autoeval,liu2026trustworthy}. This accounting prevents a better Harness from being mistaken for a better model and makes explicit which parts of the four-target program are demonstrated here.

\subsection{System Integration, Oversight, and Governance}
\label{sec:system-integration}
In physical tasks, the action model is one component in this stack. A VLA/WAM maps visual, language, and robot-state inputs to short action chunks, but it need not own task decomposition, completion detection, or recovery. The coding agent can select a camera view, call perception and navigation tools, invoke a VLA/WAM or another learned action module with a local instruction, inspect the resulting state through the world program, and decide whether to continue, interrupt, or rewrite the next subgoal. This separation makes improvements compositional: a stronger action model can be plugged in without redesigning the workflow, while a better predicate, observation abstraction, or recovery rule can improve existing action capability.

Compute allocation, model routing, quantization, caching, and hardware scheduling evolve under latency, energy, and cost constraints. Human experts provide sparse but high-value feedback: demonstrations, critiques, safety approvals, and counterexamples. Governance evolves through audit logs, identity and access control, license checks, privacy filters, release gates, and incident reviews. Because admitted traces carry the model's chain-of-thought, privacy filtering has to remove sensitive content from intermediate reasoning steps, not only from final outputs~\cite{zhou2026star}. These constraints are part of the learning system, not an afterthought.

\section{HexaAnything: A Physical Coding Agent}
\label{sec:hexanything-harness}

HexaAnything is a physical coding agent: a model that writes the world and policy programs of Section~\ref{sec:formulation}, and a Harness that runs them and establishes the coding--action--state--feedback interface. The model is a general-purpose language model or, in Section~\ref{sec:model-evolution}, HexaModel trained on traces collected by the Harness.

The Harness is the concrete boundary between the executable world and policy programs and the physical environment. It determines which observations are available, which tools can be called, how actions are monitored, what counts as evidence, and how a failure becomes a candidate revision. This section describes the runtime before the evaluation: its formal definition, verifier, staged roadmap, tool and workflow interfaces, and execution contracts.

\subsection{Formal System Definition}
\begin{figure}[!tp]
    \centering
    \includegraphics[width=0.98\linewidth]{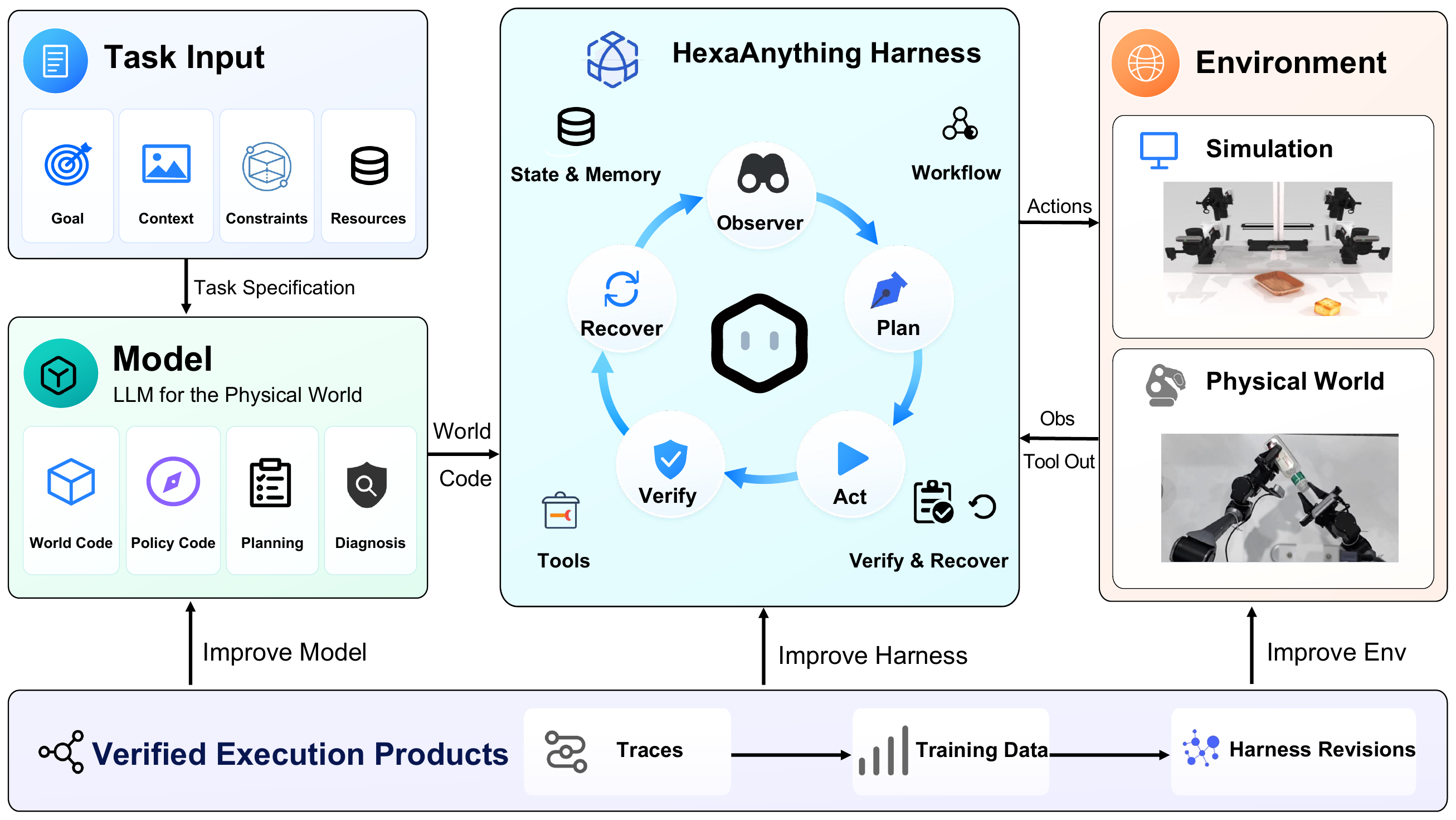}
    \caption{\textbf{Task inputs guide a physical-world model that produces world and policy code for the Harness's Observer--Plan--Act--Verify--Recover loop.} Actions
interact with simulated or physical environments, while verified traces
feed training data and subsequent model, Harness, and environment
improvement.}
    \label{fig:hexaanything-harness}
\end{figure}

We model HexaAnything as a typed runtime around two executable artifacts. Let $W_t$ be the Code-as-World program at time $t$, $P_t$ the Code-as-Policy workflow, and $H_t$ the Harness. The Harness contains an observation interface $O_t$, a tool registry $\mathcal{T}_t$, a verifier $V_t$, a recovery operator $R_t$, and a provenance-aware memory $M_t$:
\begin{equation}
    H_t = (O_t,\mathcal{T}_t,V_t,R_t,M_t,\Gamma_t),
\end{equation}
where $\Gamma_t$ denotes execution contracts, permissions, budgets, and safety gates. A coding agent with model state $\theta_t$ receives a task specification $z$ and generates a typed policy step from the current world program and observation history,
\begin{equation}
    n_t = \pi_{\theta_t}(z,W_t,P_t,O_{\leq t},M_t),
    \qquad
    u_t = \operatorname{Dispatch}(n_t,\mathcal{T}_t).
\end{equation}
The dispatched tool $u_t$ acts on environment state $x_t$ and returns an outcome $y_t$; the world program is then updated from observations and tool evidence,
\begin{equation}
    x_{t+1} \sim E(x_t,u_t),
    \qquad
    W_{t+1}=\operatorname{UpdateWorld}(W_t,O_{t+1},y_t),
    \qquad
    q_t=V_t(W_{t+1},y_t,z).
\end{equation}
Here $q_t$ is a typed verdict rather than a model-generated completion token. Depending on $q_t$, the Harness continues $P_t$, interrupts the tool, invokes $R_t$, requests a new observation, or terminates the episode. The complete trace
\begin{equation}
    \xi = \{z,W_{0:T},P_{0:T},O_{0:T},u_{0:T},y_{0:T},q_{0:T},M_{0:T}\}
\end{equation}
is the unit of attribution and later update. This definition makes explicit what HexaAnything adds around a VLA/WAM: the action model supplies one possible tool, while the Harness owns state bookkeeping, dispatch, verification, recovery, and trace admission (Figure~\ref{fig:hexaanything-harness}).

\subsection{The Verifier}
\label{sec:verifier}
The verifier is the mechanism that makes self-evolution observable and limits drift. During an episode, HexaAnything's verifier evaluates the predicates of the world program, such as whether both ingredients are inside the oven, against evidence the agent can obtain: images, depth, proprioception, and tool outcomes, and on a physical robot also sensor readings and safety limits. In simulation, the benchmark's own success check assigns the trial label; a completion claim by the model or a ``finished'' status from the action model is never counted as success (Section~\ref{sec:execution-contracts}).

Because verified traces become training data, the verifier must remain independent of the model that proposes actions; a verifier that the policy can influence would confirm its own errors. Changes to the verifier are therefore tested on a separate suite before adoption (Section~\ref{sec:evolution-space}), and disagreements between the in-episode verdict and the benchmark label are kept as counterexamples.

\subsection{Evolution Roadmap and Deployment Protocol}
\label{sec:roadmap}
This subsection operationalizes the self-evolution framework. The previous sections defined the runtime and its artifacts; here we specify when an interaction produces evidence, how a candidate update is tested, and how the program moves from Harness bootstrap toward physical deployment.
Our roadmap follows a recursive model--Harness--data loop rather than a one-time transfer from simulation to hardware:
\begin{enumerate}
\item \textbf{Stage 1: Harness bootstrap.} Use existing language and action models to establish the coding--action--state--feedback interface in a controlled simulator and selected physical tasks. The initial Harness exposes tools, state observation, verifiers, recovery, logging, and data collection.
\item \textbf{Stage 2: Model--Harness co-evolution.} Let the coding agent diagnose failures, define or revise tools, modify workflows and recovery rules, and generate candidate trajectories and evaluators in a sandbox. Real tasks and expert corrections expose capability gaps; simulation scales counterexamples and local checks. Validated traces are used for post-training, so the updated model can solve familiar tasks with fewer exploratory attempts and organize the next round of Harness improvement.
\item \textbf{Stage 3: Physical feedback and deployment.} Deploy the improved system in constrained real workflows, where engineers, users, sensors, execution logs, and safety checks provide feedback unavailable from simulator state alone. Physical failures are formalized as simulation tasks where possible and returned to the next iteration. For latency-sensitive settings, the model can generate and optimize verified policies while a real-time controller or VLA executes them, yielding a multi-timescale architecture.
\end{enumerate}
This roadmap makes the recursive claim explicit: the model is simultaneously a task solver, a Harness improver, and the eventual recipient of data produced by its own exploration. Human oversight provides initialization, external validation, and safety gates; the long-term goal is to reduce the amount of human redesign required for each new task while preserving attribution and reversibility.

For deployment, the same loop can expose robot programming, route configuration, rule authoring, and data collection through a constrained coding interface. We view this as a research question rather than a product assumption: claims should be evaluated with task completion time, rework, failure rate, safety incidents, and total cost, rather than with aggregate productivity claims alone.

\subsection{Harness Architecture for Physical Coding}
HexaAnything is the executable organization of the agent: it defines tool schemas, observation and action spaces, safety gates, test runners, data logging, and recovery procedures. We separate five interfaces that are often conflated in end-to-end action models:
\begin{enumerate}
\item \textbf{Tools} expose perception, navigation, motion primitives, gripper control, and learned action modules. A VLA or WAM is therefore an available action tool, rather than the sole locus of task intelligence.
\item \textbf{Workflow} decomposes a long-horizon instruction into observable subgoals and determines when to call, interrupt, or replace a tool.
\item \textbf{Verifier} checks semantic predicates such as object containment, contact, or switch state against visual evidence, tool outcomes, and physical sensors rather than simulator state. It must remain independent of the model's self-reported completion signal.
\item \textbf{Recovery} handles failed grasps, dropped objects, stalled action models, and changed scene state through safe stopping, retries, re-observation, and replanning.
\item \textbf{Memory} stores task-, object-, and skill-level traces with provenance and admission checks, so that useful experience can be retrieved without flooding the context with irrelevant history.
\end{enumerate}
This factorization yields a concrete interface for self-evolution: a failed episode can lead to a new tool, a revised workflow, a stronger world predicate, a recovery rule, or a memory entry. It also makes attribution more tractable because each modification has a named execution boundary. In this view, the Harness is the compiler and runtime connecting code as world to code as policy: it turns state descriptions into admissible actions and turns action outcomes back into evidence for the next edit.

\subsection{HexaAnything Execution Contracts}
\label{sec:execution-contracts}
HexaAnything makes the above interfaces operational through explicit contracts rather than an informal prompt protocol. A \emph{GoalContract} keeps the user's raw task text immutable while binding a frozen runtime task and its acceptance reference. A typed policy representation exposes a small set of admissible node kinds---observation, action, VLA invocation, judging, verification, recovery, branching, sequencing, and looping---so that proposed workflows can be statically checked before execution. An \emph{ObservationEnvelope} records the provenance of every observation, distinguishing VLA output, tool output, Harness observations, and official evaluator evidence.

The verifier returns semantic verdicts rather than a binary self-report: \textsc{Pass}, \textsc{Fail}, \textsc{InsufficientEvidence}, \textsc{Blocked}, or \textsc{SafetyStop}. A model's claim that an action is finished is therefore not success, and a low-level status such as ``finished'' is not by itself evidence that the task predicate holds. Similarly, reaching a step limit is recorded as a stopping condition rather than automatically being labeled a task failure. A \emph{TrialOutcome} records task completion, stop reason, simulator step, reward, evaluator identity, and evidence source; an \emph{ExperimentSpec} fixes the task identifier, seed set, execution mode, goal mode, step and round budgets, concurrency, and task knowledge available at batch creation.

The orchestrator runs an explicit experiment--round--trial state machine. After each tool call it applies the monitor--verify--continue/intervene decision: it may continue the workflow, interrupt the current tool, retry, re-observe, or apply a local recovery edit. Recovery is fail-closed: an unfinished or ambiguous state is not silently promoted to success, and a candidate Harness edit must pass static checks, regression tests, and the official evaluator before its trace is admitted as Physical Coding data. These contracts are the mechanism by which HexaAnything turns a flexible coding agent into a reproducible Harness and make gains attributable to workflow, verifier, recovery, model, or tool changes.

\section{Experiments: Harness, Model, and Tool}
\label{sec:preliminary-evaluation}

We evaluate three claims on RoboCasa365 and RoboDojo. With the action model held fixed, moving decomposition, verification, and recovery into the Harness improves long-horizon success (Section~\ref{sec:harness-effect}). A model trained on data returned by the Harness improves over its base model when placed back in the same Harness (Section~\ref{sec:model-evolution}). Tools improve when the agent revises them against a fixed evaluator (Section~\ref{sec:tool-evolution}). Unless noted otherwise, all conditions are evaluated on the same tasks and random seeds. Section~\ref{sec:scientific-experiments} applies the same interface outside manipulation.

\subsection{Harness Effect with a Fixed Action Model}
\label{sec:harness-effect}
RoboCasa365 measures task success on three splits: Atomic-Seen, Composite-Seen, and Composite-Unseen. The action model is XR-1~\cite{fan2025xr1}, a state-of-the-art VLA on RoboCasa365, and the native VLA and every coding agent share the same maximum number of environment steps. With XR-1 as the action tool, HexaAnything driven by GPT-5.6-Sol raises success from 34.3\% to 38.3\% on Composite-Unseen (+4.0 points), from 54.8\% to 61.5\% on Composite-Seen (+6.7 points), and from 56.6\% to 61.1\% overall (Table~\ref{tab:hexa-harness}). Codex, a general-purpose coding agent driven by the same model, reaches 59.5\% overall: it is 0.8 points ahead of HexaAnything on Atomic-Seen (81.7\%) but 1.7 points behind on Composite-Seen (59.8\%) and 4.2 points behind on Composite-Unseen (34.1\%), where it does not improve on native XR-1. The gain does not require a closed model: with Qwen3.8-27B as the planner, Composite-Unseen success is 37.3\%, 3.0 points above native XR-1 (Table~\ref{tab:model-evolution}). HexaAnything's runtime matters for the rest of this section: its typed contracts make runs reproducible and attributable (Section~\ref{sec:hexanything-harness}), its traces return as training data (Section~\ref{sec:model-evolution}), and its tools can be revised against a fixed evaluator (Section~\ref{sec:tool-evolution}).

\input{tables/hexa_harness}

\subsubsection{Long-Horizon Execution on RoboCasa365}
\label{sec:case}
On three Composite-Unseen tasks with 100 seeds each, HexaAnything raises success by 31.0, 15.0, and 17.0 points over native XR-1 (Table~\ref{tab:internal-pilot}). XR-1 is the same in both conditions; in HexaAnything it is called as a tool while GPT-5.6-Sol maintains subgoals, verifies progress, and can interrupt or revise the next call.

\input{tables/internal_pilot}

\begin{figure}[t]
\centering\scriptsize
\setlength{\tabcolsep}{1pt}\renewcommand{\arraystretch}{0.9}
\newcommand{\kb}[1]{\includegraphics[width=0.243\linewidth]{figures/kebab_case/#1.jpg}}
\begin{tabular}{@{}cccc@{}}
\multicolumn{4}{@{}p{\linewidth}@{}}{\textbf{XR-1 (native)}\hfill\textcolor{red!70!black}{\ding{55}}} \\
\kb{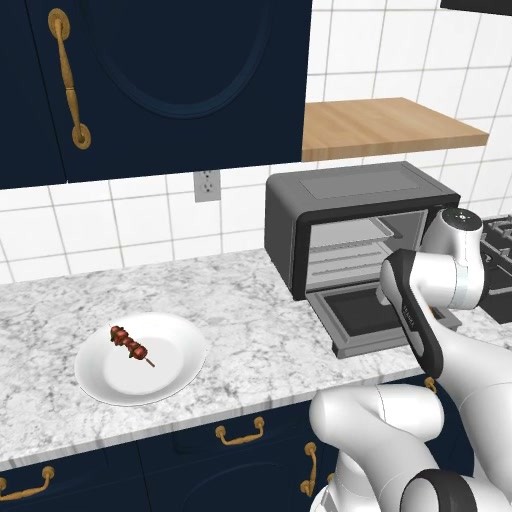} & \kb{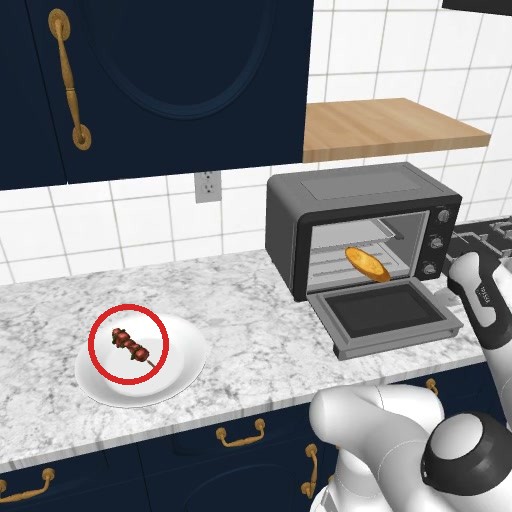} & \kb{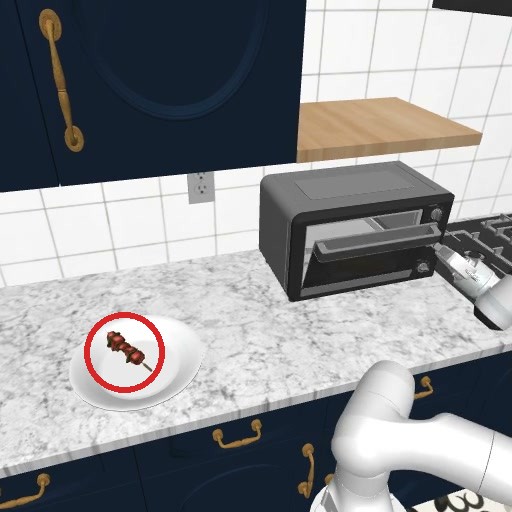} & \kb{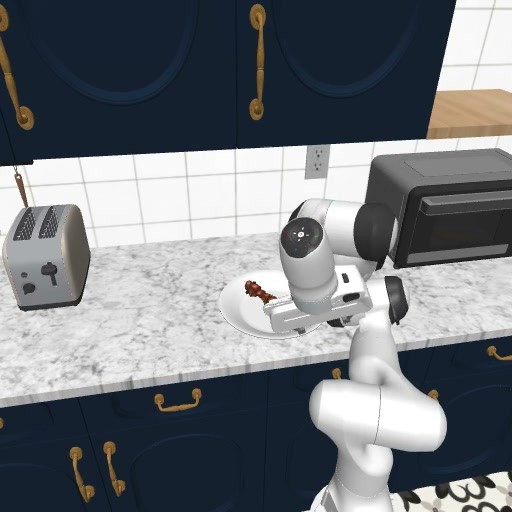} \\
\color{inkMuted}bread into oven & \color{inkMuted}kebab still on plate & \color{inkMuted}door closed early & \color{inkMuted}kebab cannot go in \\[4pt]
\multicolumn{4}{@{}p{\linewidth}@{}}{\textbf{HexaAnything}\hfill\textcolor{green!55!black}{\ding{51}}} \\
\kb{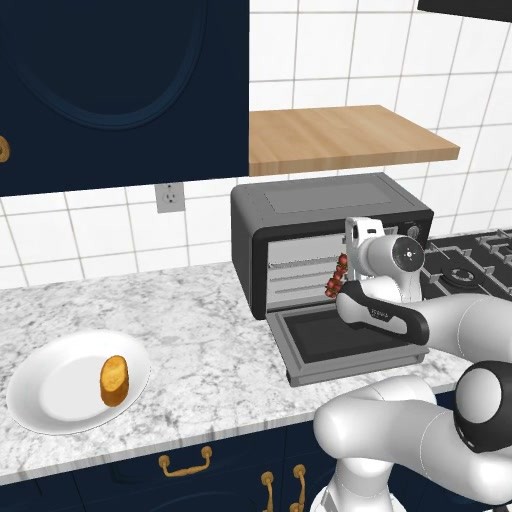} & \kb{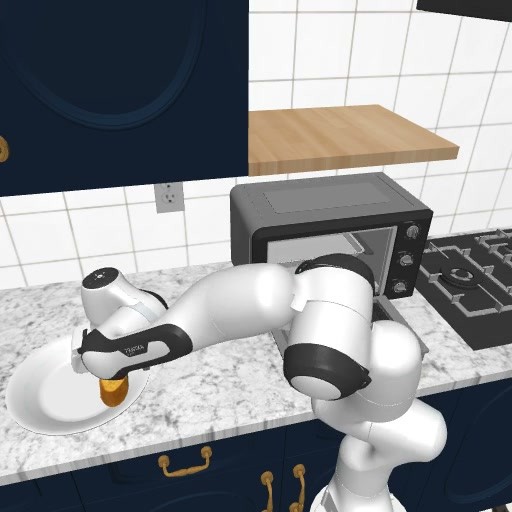} & \kb{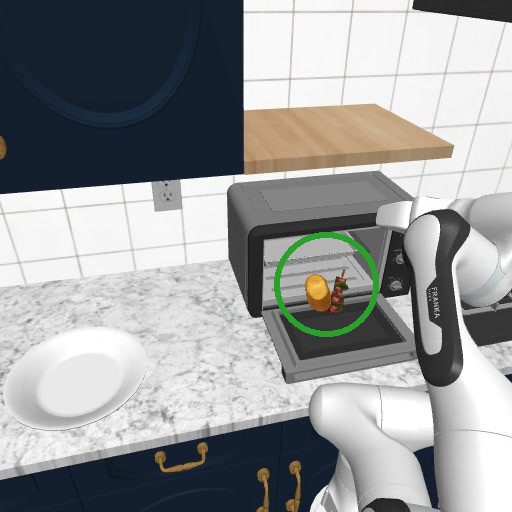} & \kb{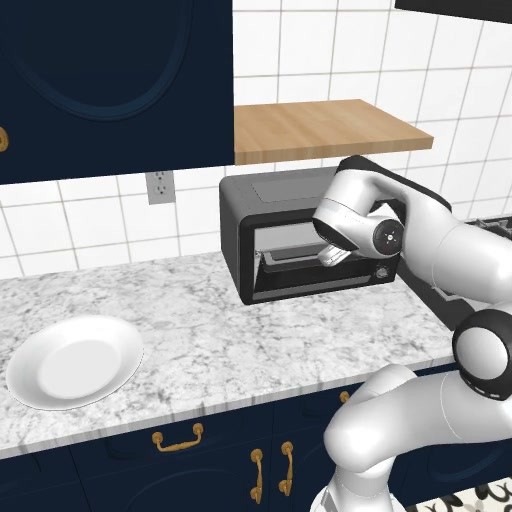} \\
\color{inkMuted}kebab into oven & \color{inkMuted}fetch bread & \color{inkMuted}both inside: close allowed & \color{inkMuted}door closed \\
\end{tabular}
\caption{\textbf{LoadKebabSandwich from the same initial state (seed 3)}. Native XR-1 puts the bread in the oven and closes the door with the kebab still on the plate (red circle); the kebab can no longer be placed. HexaAnything checks after each XR-1 call which items are inside and allows the door to close only when both are (green circle).}
\label{fig:kebab-case}
\end{figure}

The difference lies in where decisions are made. In \emph{LoadKebabSandwich}, the baseline sometimes closes the oven after placing only one ingredient, making the remaining ingredient unreachable (Figure~\ref{fig:kebab-case}). The coding agent treats ``both ingredients are inside'' as a verifier-backed precondition for closing the door; when the predicate is false, it rewrites the next subgoal and calls the action tool again. In \emph{PortionHotDogs}, native runs leave a sausage in the bowl, keep grasping inside the bowl while a plate stays empty, or stall after placing both breads; the agent instead re-estimates after each call which items remain in the bowl and which plate is missing which item, and recovers from a dropped bread, an imprecise placement, or a stalled call instead of taking the action model's termination signal as task success (Appendix~\ref{app:robocasa-cases}). In both tasks the VLA is unchanged; what changes is the workflow and the state predicate that gates it.

Context also grows with the trajectory, which success rates do not show. Long trajectories accumulate tool outputs, observation JSON, images, and polling messages; in a representative trace, replacing verbose action dumps with an on-demand execution summary reduced the active context from roughly 190K to an estimated 30--35K tokens while the full trace stayed on disk. The change is in the Harness, not the model, and it decides which evidence is available for reasoning and training.

\paragraph{Additional checks.}
A longer action budget does not explain the gain. On PortionHotDogs, doubling the VLA action budget did not improve the native condition: it completed 21 of 100 trials, compared with 22 of 100 for the standard native setting, whereas the coding-agent condition completed 37. The monitor--interrupt--replan workflow also raises CloseFridge from 50.0\% to 79.0\% and TurnOffStove from 0 of 8 baseline trials to 18.0\%; these two tasks were run under different protocols and are not pooled with Table~\ref{tab:internal-pilot}.

\subsection{Model Evolution}
\label{sec:model-evolution}
\label{sec:self-evolution}
HexaModel v0.1 is Qwen3.8-27B fine-tuned on data returned by the Harness together with general-domain data. The Harness contributes two streams (Table~\ref{tab:data-evolution}): 9.3K RoboCasa365 VQA examples with chain-of-thought, covering state and task-completion judgments, and 1.0K agent traces containing subgoals, tool calls, verifier outcomes, and recovery decisions. The traces come from three planners: 703 successful runs of the base Qwen3.8-27B, 204 recovery runs of GPT-5.6-Sol, and 103 recovery runs of an earlier fine-tuned checkpoint, so part of the training data is produced by a model trained in a previous round. Together, Harness-returned data make up 38.8\% of the 178.6M training tokens; the rest is general embodied VQA, scene captions, and general-domain math, code, and instruction data.

\input{tables/data_evolution}

\begin{figure}[t]
\centering
\begin{tikzpicture}
\begin{axis}[
    width=\linewidth, height=5.4cm,
    ybar=1.5pt, bar width=14pt,
    ymin=25, ymax=92, ytick={25,50,75},
    ylabel={Success rate (\%)}, ylabel style={font=\scriptsize, color=inkMuted},
    symbolic x coords={Atomic-Seen,Composite-Seen,Composite-Unseen,Overall},
    xtick=data, xticklabels={Atomic-Seen,Composite-Seen,Composite-Unseen,\textbf{Overall}},
    x tick label style={font=\scriptsize, yshift=-1pt},
    y tick label style={font=\scriptsize, color=inkMuted, /pgf/number format/assume math mode=true},
    enlarge x limits=0.14,
    axis x line*=bottom, axis y line=left, y axis line style={draw=none},
    x axis line style={black!55}, major tick length=0pt,
    ymajorgrids, major grid style={black!10},
    legend style={at={(0.5,1.02)}, anchor=south, legend columns=4, draw=none, font=\scriptsize, /tikz/every even column/.append style={column sep=10pt}},
    legend image code/.code={\fill[#1] (0cm,-0.08cm) rectangle (0.26cm,0.08cm);},
    nodes near coords={\pgfmathprintnumber[fixed,precision=1,fixed zerofill,assume math mode=true]{\pgfplotspointmeta}},
    every node near coord/.append style={font=\tiny, color=black!80, anchor=south, yshift=-0.5pt},
    clip=true, 
]
\fill[hexasurface] ({rel axis cs:0.79,0}) rectangle ({rel axis cs:1,1});
\addplot[fill=evoVLA, draw=none]
  coordinates {(Atomic-Seen,78.0) (Composite-Seen,54.8) (Composite-Unseen,34.3) (Overall,56.6)};
\addplot[fill=evoBase, draw=none]
  coordinates {(Atomic-Seen,80.8) (Composite-Seen,61.0) (Composite-Unseen,37.3) (Overall,60.5)};
\addplot[fill=evoGPT, draw=none]
  coordinates {(Atomic-Seen,80.9) (Composite-Seen,61.5) (Composite-Unseen,38.3) (Overall,61.1)};
\addplot[fill=evoHex, draw=none]
  coordinates {(Atomic-Seen,81.0) (Composite-Seen,62.3) (Composite-Unseen,39.5) (Overall,61.7)};
\legend{XR-1 (native{,} no Harness), Qwen3.8-27B (base), GPT-5.6-Sol, HexaModel v0.1}
\end{axis}
\end{tikzpicture}
\caption{\textbf{Success rates of Table~\ref{tab:model-evolution} by split on RoboCasa365.} XR-1 (native) runs the VLA without the Harness; the other three bars are the planner inside HexaAnything with XR-1 as the action tool. Each task is evaluated on 50 seeds (18 Atomic-Seen, 16 Composite-Seen, and 16 Composite-Unseen tasks).}
\label{fig:model-evolution}
\end{figure}
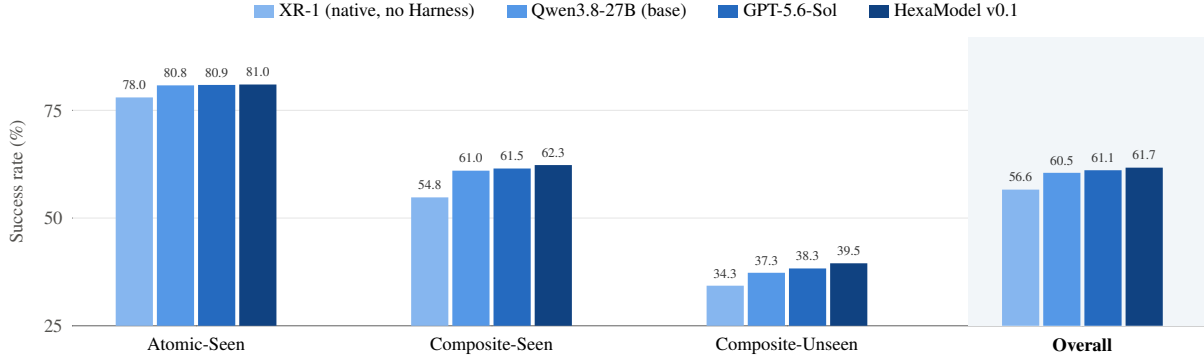

Placed back in the same Harness, HexaModel v0.1 improves over its base model on every split (Table~\ref{tab:model-evolution}, Figure~\ref{fig:model-evolution}): from 80.8\% to 81.0\% on Atomic-Seen, from 61.0\% to 62.3\% on Composite-Seen, from 37.3\% to 39.5\% on Composite-Unseen, and from 60.5\% to 61.7\% overall. The largest gain, 2.2 points is on Composite-Unseen, the split that depends most on decomposition and recovery, where the open 27B model reaches 39.5\%, 1.2 points above GPT-5.6-Sol (38.3\%) in the same Harness. Overall, HexaModel v0.1 (61.7\%) is 0.6 points above GPT-5.6-Sol (61.1\%) and ahead of it on every split, by 0.1--1.2 points.

\input{tables/model_evolution}

\subsection{Tool Evolution}
\label{sec:tool-evolution}
RoboDojo is a unified simulation and real-robot benchmark for manipulation policies~\cite{chen2026robodojo}. On three of its simulated tasks, the task definition and the evaluator stay fixed; only the tools and the workflow that calls them change. After each round a programming agent reads the failed traces, diagnoses the cause, and edits the tool code, and the revised tools are scored by the same task evaluator in the next round. Over two revision rounds, success on Fold cloth, Pour vase, and Press by number rises from 0.0\%, 40.0\%, and 0.0\% to 80.0\%, 100.0\%, and 100.0\% (Table~\ref{tab:tool-evolution}). Improvement is not monotonic: the first revision of Pour vase lowers success from 40.0\% to 0.0\%, and the second, informed by those failures, raises it to 100.0\%. The weights are unchanged throughout, so these gains come from the Harness alone. Section~\ref{sec:real-world} repeats the procedure on a physical robot.

\input{tables/tool_evolution}

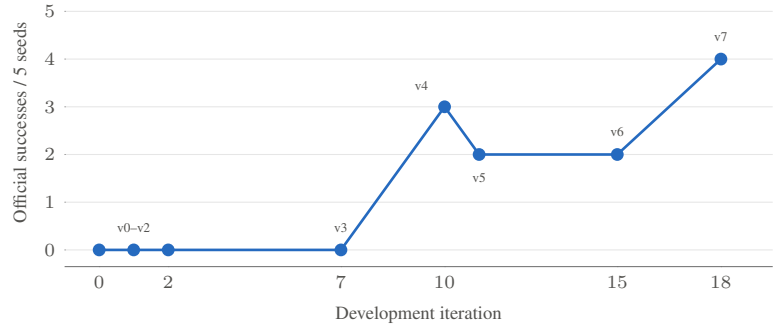
\begin{figure}[H]
\centering
\begin{tikzpicture}
\begin{axis}[
    width=0.66\linewidth, height=5.4cm,
    xlabel={Development iteration}, ylabel={Official successes / 5 seeds},
    xlabel style={font=\scriptsize, color=inkMuted}, ylabel style={font=\scriptsize, color=inkMuted},
    xmin=-1, xmax=19.5, ymin=-0.35, ymax=5.7,
    xtick={0,2,7,10,15,18}, ytick={0,1,2,3,4,5},
    tick label style={font=\scriptsize, color=inkMuted},
    axis x line*=bottom, axis y line=left, y axis line style={draw=none},
    x axis line style={black!55}, major tick length=0pt,
    ymajorgrids, major grid style={black!10},
    clip=false,
]
\addplot[color=vaDark, mark=*, mark size=1.9pt, line width=1pt]
  coordinates {(0,0) (1,0) (2,0) (7,0) (10,3) (11,2) (15,2) (18,4)};
\node[font=\tiny, color=inkMuted, anchor=south] at (axis cs:1,0.18) {v0--v2};
\node[font=\tiny, color=inkMuted, anchor=south] at (axis cs:7,0.18) {v3};
\node[font=\tiny, color=inkMuted, anchor=south east] at (axis cs:9.8,3.15) {v4};
\node[font=\tiny, color=inkMuted, anchor=north] at (axis cs:11,1.8) {v5};
\node[font=\tiny, color=inkMuted, anchor=south] at (axis cs:15,2.18) {v6};
\node[font=\tiny, color=inkMuted, anchor=south] at (axis cs:18,4.18) {v7};
\end{axis}
\end{tikzpicture}
\caption{\textbf{Tool self-evolution on RoboDojo Fold cloth, development record.} Each point is one version of the pick-and-place tool, placed at the development iteration in which it landed and replayed on seeds 0--4 with one fixed plan per seed; for three of the five seeds that plan was chosen with access to diagnostic garment keypoints, so the re-evaluation of the final version without privileged information (3/5 development seeds, 4/5 held-out seeds) is reported in the text. v0 rejects the garment as too wide; v1--v2 add the pinch mode for thin layers; v3 synchronizes the two arms; v4 tilts the wrists apart; v5 releases vertically; v6 raises the drop point over folded layers; v7 separates the hover clearance at the drop.}
\label{fig:sim-tool-evolution}
\end{figure}

Figure~\ref{fig:sim-tool-evolution} resolves the Fold cloth column of Table~\ref{tab:tool-evolution} into individual tool versions; development iterations are finer-grained than the revision rounds of the table. A long-sleeved shirt has to be folded so that both sleeves and the hem satisfy RoboDojo's evaluator. The agent did not write a folding routine: it revised one reusable two-point pick-and-place tool, which a human operator and an agent call through the same interface; an episode is three calls (left sleeve, right sleeve, two-arm body fold), each preceded by a new observation. The starting version treated the shirt as a 60\,cm-wide rigid object and rejected every call. Seven revisions, the last landing at development iteration 18, added a pinch mode for thin deformable layers, synchronized dual-arm transport, an outward wrist tilt that keeps the two wrists apart, a vertical release pose, a drop height taken from the highest folded layer, and a separate hover clearance at the drop point. Each version keeps the previous call signature and ships with unit tests; the tool code contains no branch on seed, garment, or task, and the task, the evaluator, and the weights were fixed throughout. Figure~\ref{fig:sim-tool-evolution} is the development record: each version, replayed on seeds 0--4 with one fixed plan per seed, succeeds on 0, 0, 0, 0, 3, 2, 2, and 4 of 5 episodes; the curve is not monotonic, since two revisions traded one failure for another. Because the operator had consulted the recorder's garment keypoints, which share their source with the official check, when choosing pixels for three of those seeds, the frozen final version was re-evaluated without privileged information: the keypoint output was removed from the recorder; the evaluator's source and internal state were not consulted during the evaluation (the operator had read the evaluator's source in earlier development sessions); and pixels were chosen only from the public RGB-D images, the camera calibration, and the tool's own returns, the inputs an execution agent receives. Under this constraint the final tool succeeds on 3 of the 5 development seeds and on 4 of 5 held-out seeds (5--9) that were not used to develop the tool or the pixel-selection rules, each run once in a fresh container without retries. Seed 3 fails in every version because the garment lies at the edge of the dual-arm workspace, and seed 9 completes all three folds without passing the check. The operator chose the pixels for each call, so these numbers measure the tool and the interface rather than autonomous perception. The programming agent was Claude Fable 5.1, with Claude Opus 5 completing the re-evaluation, and the whole history took one development day plus the re-evaluation.

\FloatBarrier
\section{Beyond Manipulation: Scientific Experiment with Physical Coding}
\label{sec:scientific-experiments}

\begin{figure}[t]
\centering
\includegraphics[width=\linewidth]{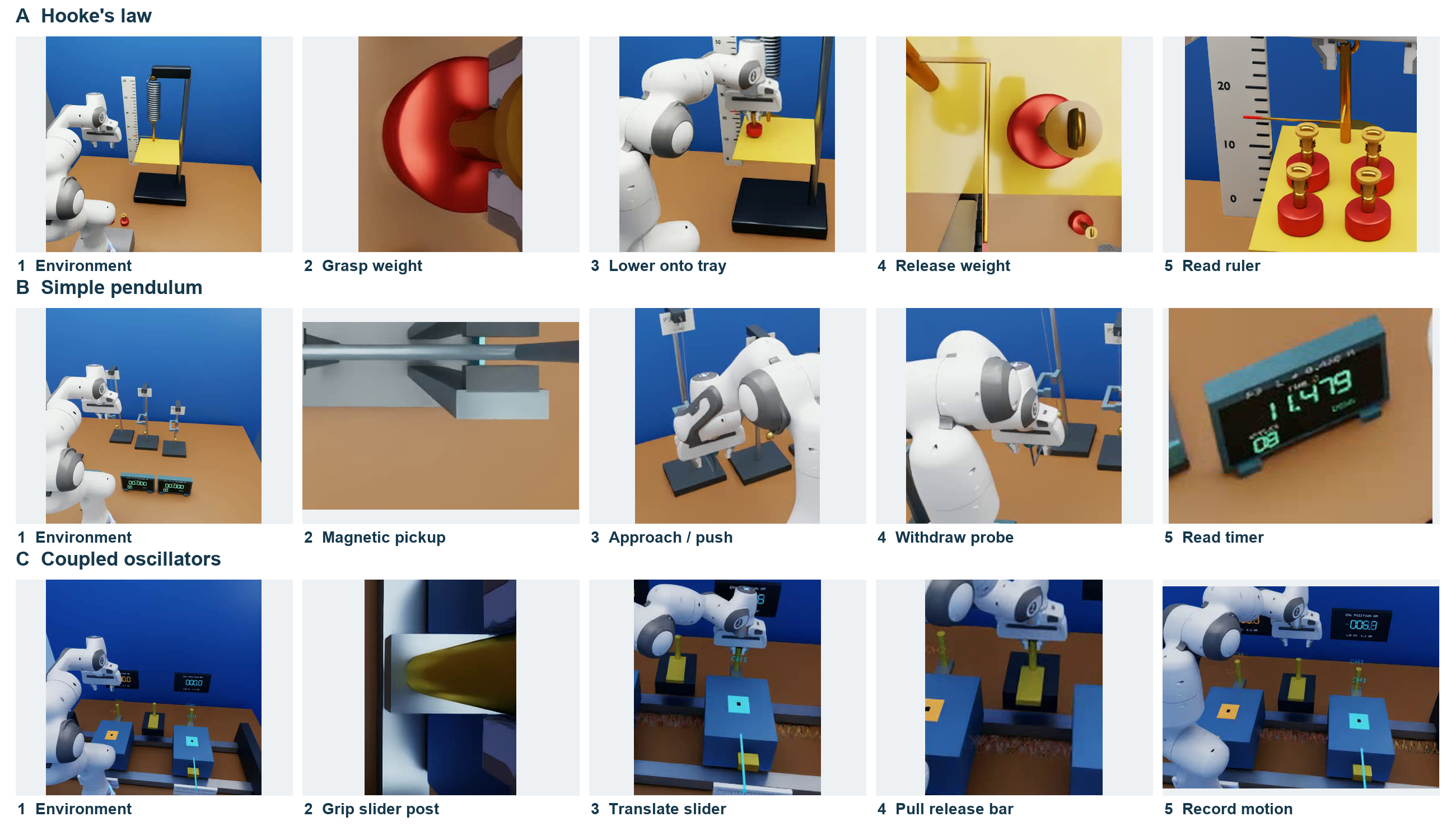}
\caption{\textbf{The three PhyBench experiments and the robot operations they require.} (A) Hooke's law: place weights on the spring-supported tray and read the ruler. (B) Simple pendulum: push a bob with a magnetic probe, withdraw, and time complete cycles. (C) Coupled oscillators: displace a slider, pull the release bar, and record both sliders. Frames are from simulation runs and show one feasible procedure; the agent designs its own.}
\label{fig:phybench-workflows}
\end{figure}

The manipulation tasks of Section~\ref{sec:preliminary-evaluation} succeed when objects reach a target arrangement; a scientific experiment succeeds only when its quantitative conclusion is correct. We develop \textbf{PhyBench}, a simulated laboratory with three tasks: estimating a spring constant, gravitational acceleration, and the normal-mode frequencies of coupled oscillators. Each task specifies the objective and the available apparatus and instruments but not the procedure; the agent must design the experiment, operate the apparatus, take measurements, and report an estimate. Figure~\ref{fig:phybench-workflows} shows the three experiments and the robot operations each requires.

In PhyBench, manipulation is only the means; the deliverable is a quantitative claim backed by evidence. A sequence of successful grasps does not by itself produce a force--extension regression, a period--length model, or a two-channel modal analysis. An agent completing these experiments must read instruments, record each measurement together with the condition that produced it, fit a model, and use the residuals to decide whether to measure again. The output of an action model---an action sequence and a stop signal---has no place for these intermediate quantities, so such tasks cannot be handed to it end to end. HexaAnything represents observation, action, and feedback uniformly as code: the agent reads measurements from sensors and records them with their experimental conditions (Code as World), invokes robot tools through code to act, and analyzes the data in code to decide its next step (Code as Policy). The path from robot interaction to quantitative claim thus lies within a single executable and inspectable workflow.

\subsection{PhyBench: A Simulated Laboratory}

\paragraph{From Isaac Sim to instrumented experiments.}
We construct the laboratory in NVIDIA Isaac Sim, using USD scene assets and PhysX-based simulation~\cite{nvidiaIsaacPhysics}. Each task combines a Franka Panda arm and parallel gripper, a tabletop apparatus, calibrated cameras, and task-specific instruments. Apparatus geometry is paired with physical constraints: a spring-supported loading tray, hinged pendulum rods, or two sliders constrained to a common rail. Task-specific dynamics and instrument logic complement the simulator's rigid-body and contact execution. Instruments use simulation time, not model or network latency. The pendulum and coupled runtimes advance physics at 240\,Hz; the coupled displacement recorder samples at 120\,Hz.

An episode seed fixes the apparatus parameters. Reference values such as the true stiffness, gravity, and eigenfrequencies are hidden from the agent and used by the evaluator only after a result is submitted, and all measurement evidence is bound to its episode.

\paragraph{Experiment interface.}
PhyBench provides the simulated apparatus, basic controls, observations, instruments, and evidence-based evaluation independently of the participating system. HexaAnything's Harness includes SDK tools for perception, motion planning, grasping, placement, and contact control, alongside terminal, file, and Python tools. Each run begins with planning. From the public task specification and an initial observation of the scene, the agent identifies the apparatus, instruments, and their labels, and decides which experimental conditions to create, how to measure each one, and how the data will be analyzed. It then carries out the plan with robot tools, checks each measurement, and revises the plan when a step fails. Scoring depends on physical outcomes and submitted evidence rather than on the use of a particular tool.

\FloatBarrier

\subsection{Three Experimental Tasks: Objectives and Apparatus}

\paragraph{Hooke's law: estimating spring stiffness.}
The environment contains a spring-supported tray, a visual ruler, and a seed-dependent set of three to five weights. The objective is to estimate the unknown spring constant from the forces and deformations the agent produces and measures. For added force $F_i=m_i g$ and extension $\Delta x_i$ (measured from the ruler in the direction of spring extension), a working model is
\begin{equation}
F_i=k\Delta x_i+b.
\end{equation}
The agent chooses the fitting convention and writes code to estimate $k$, inspect residuals, and describe uncertainty; the offset $b$ exposes an imperfect zero reference instead of hiding it in a single ratio. Camera images can be cropped or enlarged on demand, but the benchmark does not provide the correct reading. A valid run must physically support and release the weights, cover the required loading conditions, obtain stable readings, and submit evidence within budget. Success requires a relative error in $k$ of at most 15\%.

\paragraph{Simple pendulum: estimating gravitational acceleration.}
The objective is to infer the unknown gravitational acceleration using three labeled pendulums, P1--P3, with different publicly calibrated effective lengths. The apparatus provides gold bobs, fixed supports, a magnetically attachable probe, and optical gates that measure complete free-swing cycles. The agent must identify the relevant objects and instruments, decide how to excite the pendulums within their swing planes, and obtain timing evidence without continued robot contact. For $N_i$ cycles measured over $\tau_i$, the period is $T_i=\tau_i/N_i$, and the small-angle relationship
\begin{equation}
T_i^2=\frac{4\pi^2}{g}L_i
\end{equation}
allows $g$ to be estimated from periods measured at different lengths. Valid timing requires complete cycles after warm-up, free motion during timing, and an amplitude below $5^\circ$; blocked instruments, incomplete cycles, or robot contact invalidate the evidence. Success requires a relative error in $g$ of at most 3\%.

\paragraph{Coupled oscillators: identifying normal-mode frequencies.}
The objective is to identify two normal-mode frequencies from the motion of two spring-coupled sliders. Available components include a common rail, individual preparation brakes, CH1/CH2 grip posts, a physical release bar that opens both brakes, and a dual-channel displacement recorder. The agent designs initial conditions using these components and determines how to excite and observe the system. The recorder exports synchronized $x_1(t),x_2(t)$ as CSV at 120\,Hz and 0.1\,mm resolution, so the time series need not be reconstructed from images. The agent can write Python to fit a two-mode model,
\begin{equation}
x_c(t)=a_c+\sum_{j=1}^{2}e^{-\gamma t}
\left[A_{cj}\cos(\omega_{d,j}t)+B_{cj}\sin(\omega_{d,j}t)\right],\quad c\in\{1,2\},
\end{equation}
and recover $\omega_j=\sqrt{\omega_{d,j}^2+\gamma^2}$ under the public proportional-damping convention. Valid evidence requires two independent initial displacements, free recordings of at least 40\,s with measurable motion on both channels, and budget compliance. Success requires a relative error of at most 10\% for each frequency.

\subsection{Execution Protocol and Results}

An autonomous run receives the public task together with HexaAnything's experiment guidance, a participant-side prompt that recommends general measurement practices such as recording each measurement immediately, interpolating between visible scale marks, and fitting in code. The run then proceeds without intervention under a 7,200\,s wall-clock limit and an 18,000-frame simulation budget; the model chooses its procedure, observations, code, and final estimate.

A run is \emph{valid} if it completes the physical and evidence protocol and receives a finite error, even if that error exceeds the success threshold. For $m$ target quantities, the reporting metric is
\begin{equation}
\mathrm{MRE}=\frac{100\%}{m}\sum_{j=1}^{m}
\frac{|\hat q_j-q_j^{\mathrm{ref}}|}{|q_j^{\mathrm{ref}}|}.
\label{eq:scientific-mre}
\end{equation}
Table entries average this per-run error over valid runs and are reported together with the valid-run count $n/N$. Failed or interrupted trials remain in $N$ without being assigned a numerical error.

\begin{figure}[t]
\centering
\includegraphics[width=\linewidth]{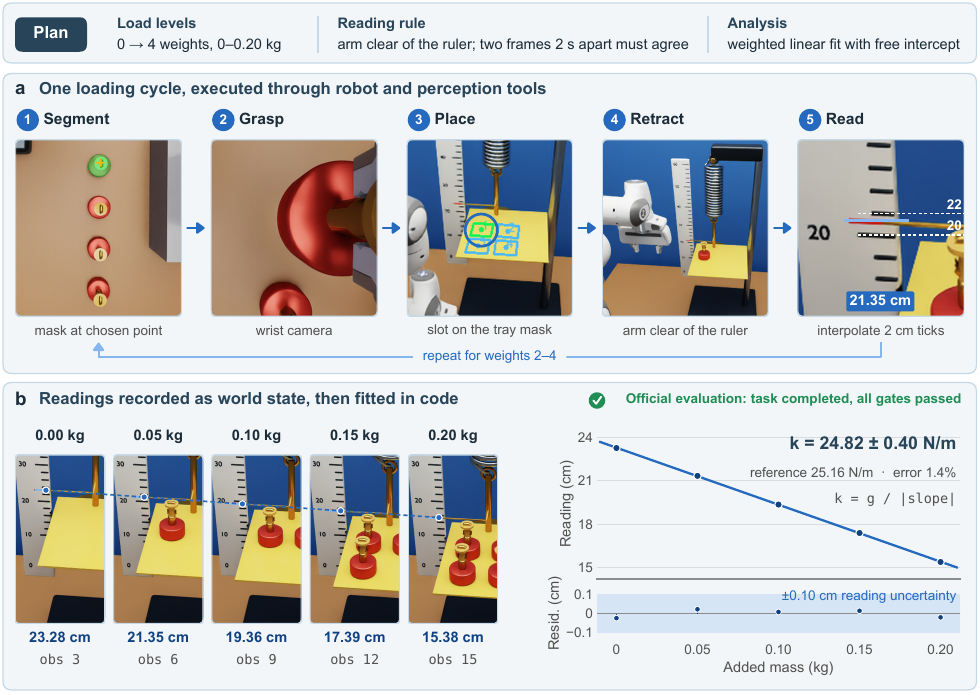}
\caption{\textbf{A Hooke's-law run with HexaAnything and Opus 5.5.} One of the ten runs in Table~\ref{tab:scientific-experiments}; the plan (top) precedes any action. (a) One loading cycle, repeated for each of the four weights: segment, grasp, place on the tray, move clear of the ruler, and read the pointer between the 22 and 20\,cm ticks. (b) Readings at the five load levels, each with its source observation (obs~$n$), and the weighted linear fit: $k=24.82\pm0.40$\,N/m against a reference of 25.16\,N/m (error 1.4\%). Masks and slot grids are tool outputs; circles, dots, and lines are added.}
\label{fig:hooke-case-main}
\end{figure}

\input{tables/sci_expriment}

\paragraph{Autonomous execution.}
With GPT-6-Astra, Opus 5.5, and GPT-5.6-Sol, HexaAnything completes all three experimental end to end (Table~\ref{tab:scientific-experiments}), from designing the procedure and manipulating the apparatus to collecting measurements, analyzing data, and submitting results. GPT-6-Astra and Opus 5.5 obtain valid results in all 30 trials each, and GPT-5.6-Sol in 27 of 30. All three models keep the mean relative error below 5\% on every task: Opus 5.5 is most accurate on Hooke's law (1.3\%) and the simple pendulum (1.0\%), and GPT-6-Astra on the coupled oscillators (0.8\%). Qwen3.8-Max obtains valid estimates in only 5 of 10 Hooke's-law trials and 3 of 10 trials on each of the other two tasks, with mean errors of 12.4\%, 2.4\%, and 3.7\%. Both completion and measurement accuracy therefore depend on the underlying model.

\subsection{Case Study: Hooke's Law}
\label{sec:hooke-case}

We follow one of the ten Hooke's-law runs with Opus 5.5 from plan to estimate (Figure~\ref{fig:hooke-case-main}); example runs of the other two tasks, shown in the same form, are given in Appendix~\ref{app:phybench-cases}. The run was selected because it completed the official evaluation with all gates passed and kept a complete trace; its relative error in $k$ of 1.36\% is close to this model's mean of 1.3\% on the task.

\paragraph{Plan.}
After observing the scene---a tray hanging from the spring, a pointer in front of the ruler, and four weights on the table---the agent planned to record the unloaded reading as the zero reference, add the weights one at a time to obtain five load levels from 0 to 0.20\,kg, take each reading only after moving the arm away from the ruler and confirming that two images taken 2\,s apart agree, and fit a linear model with a free intercept.

\paragraph{Execution.}
For each weight, the agent segmented it at a point it selected, grasped it, placed it in a free slot on the segmented tray, and moved the arm clear of the ruler (Figure~\ref{fig:hooke-case-main}a). It read the pointer at native image resolution by locating the pointer and the adjacent 2\,cm ticks and interpolating between them, at about 0.11\,cm per pixel with a standard uncertainty of 0.10\,cm. Each reading was recorded in the world state together with the load that produced it and the ID of the observation it came from (Figure~\ref{fig:hooke-case-main}b).

\paragraph{Analysis.}
A weighted least-squares fit with a free intercept gives $k=24.82\pm0.40$\,N/m against a reference of 25.16\,N/m, which the evaluator reveals only after submission: a relative error of 1.36\%. The largest residual, 0.024\,cm, is well below the reading uncertainty. Because the readings, their sources, and the fit are all kept as code, the reported $k$ can be traced back to five camera frames.


\FloatBarrier

\section{Physical Deployment, Limitations, and Open Problems}
\subsection{Real-World Transfer and Deployment Status}
\label{sec:real-world}

\begin{figure}[t]
\centering
\includegraphics[width=0.98\linewidth]{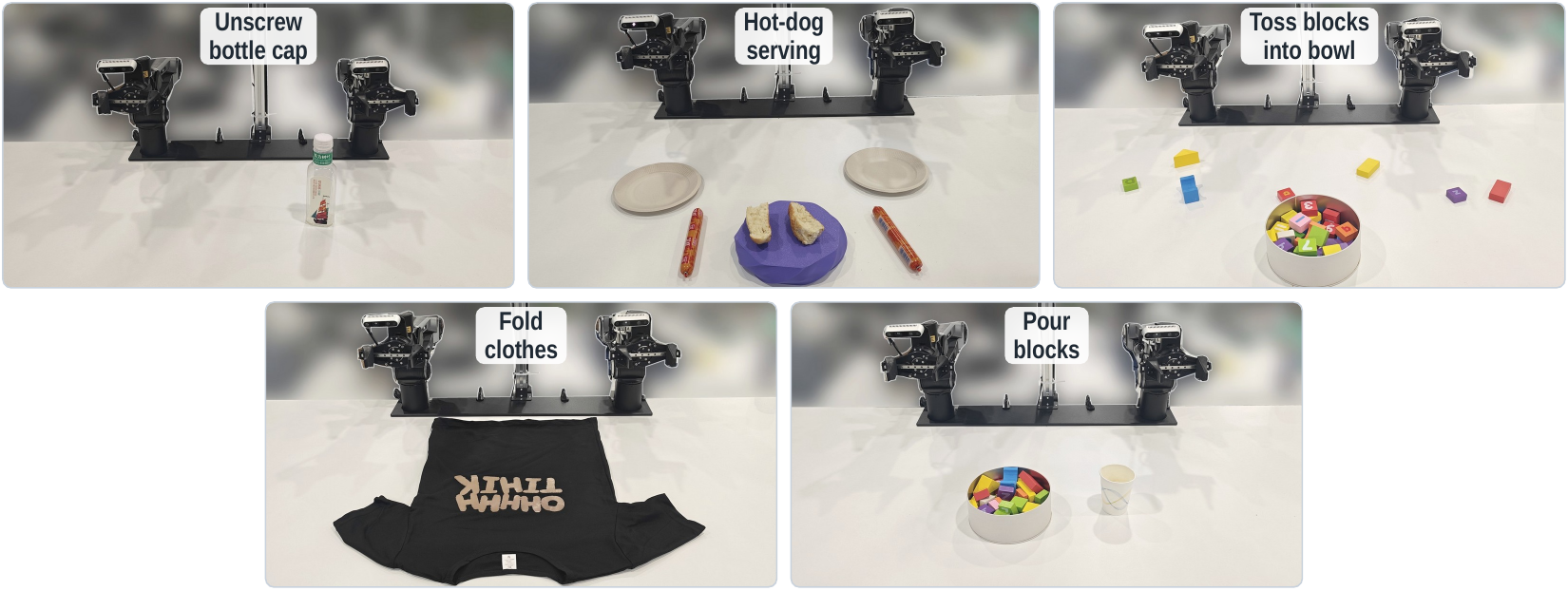}
\caption{\textbf{Initial scenes of five of the seven real-robot tasks, seen from the front scene camera:} unscrew bottle cap, hot-dog serving, toss blocks into bowl, fold clothes, and pour blocks. Tic-tac-toe and block into bowl are not pictured.}
\label{fig:real-robot-tasks}
\end{figure}

The physical platform is an AgileX PiPER-X dual-arm system: two 6-DoF arms with parallel-jaw grippers on a shared base plate, four Intel RealSense D435 RGB-D cameras (one on each wrist, two fixed on the scene), and joint-position control over CAN. On this platform HexaAnything's Harness is exposed through URAI (Universal Robot--Agent Interface), a tool collection that humans and agents share: a human draws strokes on the camera image in a browser, an agent sends the same pixels and parameters through an HTTP API, and both run through the same planning and control code on the robot host. Perception models run on off-board GPUs and the foundation model is reached through its API; the model acts only between tool calls, and each tool executes at the robot's own control rate. We call this arrangement \emph{Vibe as Policy}: a programming agent writes, validates, and freezes the tools between episodes, and a frozen execution agent composes them at run time. Seven tabletop tasks were run with HexaAnything, with GPT-6-Astra as the execution agent (Figure~\ref{fig:real-robot-tasks}, Table~\ref{tab:real-robot-tasks}).

\input{tables/real_robot_tasks}

Five of the seven tasks succeed in all three trials. Where a published result exists for the same model, HexaAnything, calling URAI tools, finishes a tic-tac-toe game in 5.0 minutes against GPT-Policy's 13.6 (2.7$\times$ faster), unscrews the bottle cap in 5.1 minutes against 17.9 (3.5$\times$), and places a block in the bowl in 1.0 minute and 443 output tokens against Robocurve's 2.5 minutes and 2.1k tokens under direct end-effector control (2.4$\times$ faster with 4.7$\times$ fewer tokens). These references come from other hardware and operating limits, and three trials per task are few. The two tasks that do not always complete fail on perception and reach rather than on the tools: in toss blocks, the last block of one round lay beside an arm base where no reliable depth could be obtained, and in fold clothes two trials stopped at 75\% completion.

\paragraph{Tool evolution on the real robot.}
Tools evolve on the robot by the loop of Section~\ref{sec:tool-evolution}, with the operator's judgment and the robot's own execution traces (measured joint speeds, gripper state, where the block landed) as evidence in place of a simulator evaluator; nothing privileged is available on the robot, since the evidence is what the cameras and joint traces record and what the operator sees. The toss tool of Table~\ref{tab:real-robot-tasks} is one such history. The agent's first versions swung the arm about its base joint and opened the gripper on a timer; the traces showed the arm trailing the reference by a quarter second at the firmware's joint-speed limit, so the release was moved to the measured arm position. The operator then recorded a single video of a human throw, and the agent wrote an overhand version from it: a wind-up folded low in front of the base, then the three pitch joints thrusting forward together. Its first trial hit a wrist joint limit during the lift. Later versions carried the block to the wind-up in joint space, ran two full-speed lines for near and far targets and chose the release instant from the landing distance, kept the reference running past the release so that the gripper opens while the arm is still at speed, and allowed the grasp to be drawn as a line and tilted toward the base for blocks lying beside the arm, where a straight-down grasp has no reliable depth. Two versions were rejected on the robot and rolled back: a faster reference that the firmware could not follow, and a high lob that the operator judged worse than the low thrust. Each version was tried on the robot within the hour and committed to the tool repository; the programming agent was Claude Fable 5.1, with Claude Opus 5 for part of one session.


\subsection{Long-Horizon Complex Tasks}
\label{sec:long-horizon-tasks}
Section~\ref{sec:harness-effect} evaluates long-horizon execution only on RoboCasa365's built-in composite tasks: the Composite-Seen and Composite-Unseen splits, including the three tasks of Table~\ref{tab:internal-pilot}. To test decomposition, search, and recovery over much longer horizons, we design a set of long-horizon tasks on RoboCasa365. Each task composes many atomic tasks into a single instruction and is substantially longer than the benchmark's composite tasks, so the agent must decompose it into subgoals itself. The robot starts at a random location in the kitchen rather than at a fixed position, and the objects a task needs are not necessarily in view: the agent has to find them by acting, for example by opening drawers. Each episode is capped at 10,000 environment steps. We have designed these tasks but not yet evaluated any system on them.

\subsection{Limitations and Open Problems}
\label{sec:limitations}
The evidence in this report is preliminary. The Harness effect is measured with one action model, XR-1, on one benchmark, and HexaAnything's overall margin over Codex driven by the same model is 1.6 points, with Codex ahead on Atomic-Seen (Table~\ref{tab:hexa-harness}). Long-horizon evidence comes from RoboCasa365's built-in composite tasks; Section~\ref{sec:long-horizon-tasks} describes longer tasks designed to extend it. The RoboDojo tool-evolution rates rest on few seeds (five for Fold cloth), and on Fold cloth a separate programming agent revised the tool while the operator chose the pixels for each call. PhyBench runs in simulation with ten trials per model and task, and the real-robot results rest on three trials per task, with published references obtained on other hardware.

Several problems remain open: preventing the model and the verifier from co-adapting, learning physical causality from sparse trajectories, keeping lifelong memories private, and combining expert corrections, simulator traces, and real-robot failures without letting the agent optimize toward a narrow or self-generated evaluator. Deployment also requires safety tests for unsafe robot actions, destructive commands, prompt injection, and secret exfiltration.

\section{Conclusion}
Physical coding agents maintain two coupled executable artifacts: a world program that records task-relevant state and a policy program that organizes planning, action, verification, and recovery. This interface makes physical execution inspectable and turns experience into persistent evidence. HexaAnything realizes this interface as a physical coding agent that observes, acts, and receives feedback through code. On RoboCasa365 it raises Composite-Unseen success from 34.3\% to 38.3\% over the native XR-1 VLA; traces it collects train HexaModel, which improves over its base model in the same Harness; and revising tools against a fixed evaluator raises success on three RoboDojo tasks. The same interface extends beyond manipulation: on PhyBench, HexaAnything autonomously designs and carries out simulated physics experiments and reports physical parameters with mean relative errors below 5\%, and on a dual-arm AgileX robot it completes five of seven tabletop tasks in all three trials. These results provide initial evidence of data, model, and tool evolution through one executable interface, but they do not establish autonomous model architecture search, unrestricted self-rewriting, or deployment in unconstrained physical environments.

The broader program has four coupled targets. Harness evolution revises tools, workflows, world representations, verifiers, and recovery. Data and environment evolution expands the experiences, tasks, objectives, and evaluators that expose capability gaps. Model evolution uses independently verified records to internalize planning, tool use, state tracking, and recovery, and eventually to search over training procedures, parameters, and architectures. Embodiment and compute evolution extends the same evaluation-gated process to sensors, robot hardware, accelerators, memory, and resource allocation. Representation and interface evolution cuts across all four: the system may eventually discover better task schemas, Physical Coding language constructs, and machine-checkable standards for aligning perception, state, and action.

Progress along this roadmap requires separate attribution. A larger corpus is not a better model, a better Harness is not a new capability in the weights, and a higher score under a changed evaluator is not evidence of generalization. Future iterations should therefore keep versioned parents, independent hold-out evaluators, protected regression suites, provenance, canary deployment, and rollback. Simulation can provide cheap resets and counterexamples; constrained physical workflows can reveal noise, latency, contact variation, and embodiment effects that simulation omits. Together they provide the feedback needed to move from Harness bootstrap toward reliable self-evolution in manufacturing and scientific workflows.

{
\small
\bibliographystyle{plainnat}
\bibliography{refs}
}

\clearpage
\appendix
\section{Extended Related Work}
\label{app:related-work}
This appendix gives the extended review of self-evolving systems that Section~\ref{sec:rw-self-evolving} summarizes, organized by the object of adaptation and the boundary within which other components are held fixed.

\subsection{Evolving Data, Tasks, Objectives, and Evaluators}

A substantial body of work shifts adaptation upstream from the policy to the data and task distributions on which policies are trained and evaluated. GenSim and RoboGen use language models to generate executable robotic tasks, scenes, and supervision, making the task-generating process itself programmable rather than fixed~\cite{wang2024gensim,wang2023robogen}. Holodeck, SAGE, and SceneSmith extend this idea to simulator-ready environments, where language-conditioned generation or refinement changes the distribution of scenes presented to downstream policies~\cite{yang2024holodeck,xia2026sage,pfaff2026scenesmith}. CurricuLLM introduces adaptation at the curriculum level, using policy performance to modify the sequence or difficulty of training tasks~\cite{ryu2025curricullm}. In contrast, RoboCasa provides a large but fixed simulation and task substrate against which policies can be trained and evaluated~\cite{nasiriany2024robocasa}. These distinctions are important because task generation, curriculum adaptation, and policy adaptation operate on different objects even when they appear in a common training loop.

Related work also treats demonstrations and trajectories as evolving artifacts. GenSim2, MimicGen, RoboTwin, and HumanoidGen synthesize executable demonstrations or trajectories that are subsequently used to train new policies~\cite{hua2024gensim2,mandlekar2023mimicgen,mu2025robotwin,jing2026humanoidgen}. The primary adaptive object in these systems is therefore the data distribution $D$, while the downstream policy-learning procedure is typically externally specified. In software engineering, SWE-bench provides a complementary fixed setting: repository-level issues and test suites define a stable task substrate for evaluating coding agents~\cite{jimenez2024swe}. Together, these works demonstrate that modifying the data- and task-generating substrate can materially affect downstream capability, but they do not by themselves establish that the agent generating these artifacts has improved.

Training objectives constitute a further programmable component. Eureka and Text2Reward synthesize executable reward functions from natural-language specifications and use rollout feedback to evaluate or refine the resulting reward code~\cite{ma2024eureka,xie2024text2reward}. DrEureka extends this formulation to jointly search reward functions and domain-randomization distributions, while REvolve incorporates human feedback into reward refinement~\cite{ma2024dreureka,hazra2025revolve}. These systems establish that the optimization objective itself can be placed inside a code-generation and execution loop. However, a training reward is not equivalent to an independent verifier. When the policy, reward generator, and scorer share data or are optimized jointly, apparent gains may reflect reward exploitation or evaluator--policy co-adaptation rather than transferable capability.

Several systems move closer to explicit evaluation infrastructure. RoboPlayground compiles natural-language instructions into reproducible task specifications with assets, initialization distributions, and success predicates; AutoEval automates real-world evaluation through scene resets and success detection; and Eval-Actions introduces process-level evidence about execution quality rather than relying solely on terminal success~\cite{wang2026roboplayground,zhou2025autoeval,liu2026trustworthy}. SimFoundry further couples generated simulator scenes to policy learning and evaluation~\cite{ranawaka2026simfoundry}. These works motivate treating evaluators and verifiers as first-class programmable objects. They also expose a central difficulty for persistent self-evolution: once the task generator, learner, and evaluator are all adaptive, improvement can no longer be attributed from scalar performance alone. Independent hold-out evaluators, disagreement across evaluators, provenance tracking, and evaluation under distribution shift become necessary to distinguish capability improvement from adaptation to the measurement process.

This line of work therefore broadens the object of evolution beyond the policy. Our formulation extends this view by treating the task specification $z$, data $D$, and verifier $\phi$ as distinct but interacting components. The distinction matters because generating harder tasks, improving training rewards, and improving independent validation have different causal roles in the learning loop.

\subsection{Evolving Programs, Tools, Harnesses, and Persistent Memory}

A second line of work concerns the executable mechanisms through which agents act. SayCan and ProgPrompt ground language-level plans in externally provided skills and situated constraints, thereby restricting language generation to executable action spaces~\cite{ahn2022can,singh2022progprompt}. Code as Policies, RoboCodeX, RoboScript, RoboCoder, and CaP-X go further by representing behavior directly as inspectable executable programs~\cite{liang2023code,mu2024robocodex,chen2024roboscript,li2024robocoder,fu2026cap}. In these systems, code is not merely an output format; it serves as an intermediate representation for composing perception, control, and task logic.

Related software-agent work demonstrates that capability also depends strongly on the surrounding execution interface. ReAct, SWE-agent, and OpenHands expose explicit tools, computer interfaces, or repository operations that structure reasoning and execution~\cite{yao2023react,yang2024swe,wang2025openhands}. These systems make clear that agent behavior is jointly determined by the underlying model and by the interface through which observations, tools, and execution feedback are organized.

Embodied-agent systems increasingly make this surrounding organization an explicit object of adaptation. Thea and Guava expose robotic capabilities as callable tools while maintaining multimodal execution context; Harness VLA, SHAPER, and Zetta introduce or revise memory, critics, recovery mechanisms, skills, or other Harness components while leaving much of the base policy fixed~\cite{wang2026towards,liu2026guava,zhang2026harness,wang2026self,ding2026zetta}. Guava additionally includes model adaptation, illustrating that Harness changes and parameter updates can coexist within the same system. The Model Hardware Standard generalizes the interface perspective toward physical instruments by defining model-agnostic drivers and standardized read/write primitives~\cite{anthropic2026mhs}. Although such interface standardization is not itself evidence of self-evolution, it highlights the role of the executable boundary between models and physical systems.

These works motivate a distinction between persistent memory $M$ and the Harness $H$. Memory denotes skills, traces, and other persistent content that can be retrieved across episodes, whereas the Harness specifies how observations, tools, memory, verification procedures, and recovery logic are composed during execution. This distinction is consequential for attribution. A new skill stored in memory does not imply that model parameters have changed; a new tool-routing policy does not imply architectural adaptation; and improved task completion after a Harness revision does not, by itself, demonstrate transfer to a new interface.

Voyager provides a representative example of persistent improvement through external skills rather than parameter updates. It stores successful behaviors in an executable skill library and retrieves them for subsequent tasks, enabling cross-episode accumulation without fine-tuning the underlying foundation model~\cite{wang2023voyager}. MEMENTO and Playful Agentic Robot Learning similarly emphasize persistent skill or experience accumulation while leaving the base model largely unchanged~\cite{sygkounas2026memento,zhang2026playful}. Such systems provide strong evidence for memory- and skill-level evolution, but they should not be conflated with model-level evolution.

Taken together, these studies establish that programs, tools, Harnesses, and memory can all support persistent capability improvement. Their broader implication is that self-evolution need not begin with parameter updates. At the same time, most systems hold the programming language, tool semantics, permissions, compiler, evaluator, or base-model weights fixed. Consequently, they primarily demonstrate adaptation within a predefined executable interface rather than evolution of the complete agent--environment boundary.

\subsection{Evolving Model Parameters, Architectures, and Research Procedures}

Model-side evolution changes the mechanism that produces policies or executable artifacts rather than only modifying those artifacts externally. It is useful to distinguish learned parameters or adapters $w$, structural configuration $A$, and persistent memory $M$, with model state written as $\theta=(A,w)$. This distinction prevents parameter adaptation, architecture search, and external-memory accumulation from being treated as interchangeable forms of model evolution.

Parameter-efficient adaptation provides a generic mechanism for modifying model behavior. LoRA, for example, freezes the base model while training low-rank adapters~\cite{hu2021lora}. Such methods demonstrate efficient parameter updates, but not autonomous self-evolution unless the data selection, update trigger, training procedure, and validation loop are themselves controlled by the agent. Agent-Driven Autonomous RL and ENPIRE move closer to this setting by allowing high-level directives or agent-generated code to influence training, reward, or policy-update procedures, and by using rollout evidence to retrain downstream policies~\cite{khandelwal2026agent,xiao2026enpire}. Even in these systems, however, it is important to distinguish between an agent that modifies a downstream policy and an agent that updates its own underlying parameters.

Algorithm and program search provide a related but distinct precedent. AutoML-Zero evolves learning algorithms from primitive operations~\cite{real2020automl}. The AI Scientist automates parts of the scientific workflow, including experiment design, implementation, and analysis~\cite{lu2024ai}. AlphaEvolve and the Darwin G\"odel Machine search over executable program or agent variants, using observed execution outcomes to retain or reject candidate modifications~\cite{alphaevolve,zhang2026darwin}. These systems demonstrate that optimization procedures and agent code can themselves become search spaces, but they do not necessarily imply modification of the neural architecture of a deployed embodied model.

Autoresearch provides a particularly useful boundary case. Karpathy's formulation restricts an agent to modifying a training script, thereby allowing bounded changes to architecture, hyperparameters, optimization, and the training loop while keeping the data-preparation script, runtime budget, and evaluation protocol fixed~\cite{karpathy2026autoresearch}. The explicit restriction is methodologically valuable because it makes attribution tractable: performance changes can be associated with edits inside a controlled search boundary. At the same time, it illustrates why autoresearch is better viewed as a \textbf{search setting} than as a component category. Whether a system performs parameter evolution, architecture evolution, reward search, or Harness adaptation depends on the artifact that the search procedure is permitted to modify.

This distinction is particularly important for claims of joint model evolution. Long-term adaptation of $(A,w,M)$ raises unresolved issues that are largely absent when only one component is updated: catastrophic forgetting, regression across tasks, causal credit assignment across modules, and safe rollback after harmful updates. Existing work provides important precedents for each component, but evidence for reliable joint evolution remains limited.

\subsection{Evolving Runtimes, Simulators, and Physical Execution Substrates}

Most work on self-evolving coding agents operates within software environments in which executable feedback is dense, reproducible, and inexpensive. Repository state, compilation, unit tests, static analysis, and version control provide unusually strong supervision for repeated adaptation. This setting is therefore well suited to studying persistent self-improvement, but it also imposes an implicit boundary: the execution substrate itself is usually assumed to be fixed.

Several lines of work relax this assumption. CompilerGym and MLGO formulate compiler decisions as optimizable components rather than immutable heuristics~\cite{cummins2022compilergym,trofin2021mlgo}. KernelBench evaluates generated GPU kernels for both correctness and runtime performance, while CUDA Agent and KernelFoundry use execution and profiling feedback to search over CUDA implementations~\cite{ouyang2025kernelbench,dai2026cuda,wiedemann2026kernelfoundry}. These systems demonstrate that executable feedback can drive adaptation below the policy layer, including compiler behavior and low-level implementation choices. However, workloads, evaluation protocols, and hardware configurations are still typically held fixed, so improvements do not by themselves establish generalization across execution substrates.

Embodied systems introduce a more substantial extension because the environment is no longer fully specified by software state. Holodeck, SceneSmith, and SimFoundry generate simulator scenes, whereas RoboCasa provides a large fixed simulation substrate~\cite{yang2024holodeck,pfaff2026scenesmith,ranawaka2026simfoundry,nasiriany2024robocasa}. Evolution Gym offers an adjacent example in which robot morphology and control are searched jointly~\cite{bhatia2021evolution}. These systems broaden the space of modifiable artifacts beyond agent code, but simulation remains an incomplete proxy for physical deployment.

Real-world execution introduces sensor noise, calibration error, unmodeled dynamics, latency, contact variation, hardware wear, morphology, and safety constraints. These factors are not reducible to repository state or simulator success. Consequently, a program that executes successfully, a generated task that is solvable, and a policy that improves in simulation provide progressively stronger but still distinct forms of evidence. None alone establishes that the improvement will survive distribution shift or transfer to a physical robot.

This motivates treating simulation as a controlled \textbf{cold-start substrate} rather than the terminal environment for self-evolution. Simulation supports inexpensive generation of tasks, failures, trajectories, and candidate Harness modifications. Controlled physical deployment then provides additional evidence about causal factors that are not represented faithfully in simulation. Under this view, environment evolution encompasses not only repository context or simulator configuration but also runtime settings, sensors, robot dynamics, hardware interfaces, and eventually selected aspects of the physical execution substrate.

Table~\ref{tab:related-work} summarizes these distinctions at the mechanism level. It groups representative systems by the artifact whose adaptation is central, and compares persistence, the boundary held fixed for attribution, and the evidence used to support improvement. The table is a scope map rather than a performance ranking: the cited systems use different tasks, embodiments, and evaluation protocols.
\input{tables/related_work}

\subsection{From Component-Local Adaptation to Coupled Self-Evolution}

Across these literatures, there is now substantial evidence that individual components of an agentic system can be adapted effectively. Tasks and curricula can be generated, reward functions can be searched, evaluators can be automated, skills can accumulate in memory, Harnesses can be revised, model parameters can be retrained, training procedures can be searched, and low-level execution code can be optimized. The principal open problem is therefore not whether another individual component can be made adaptive.

What remains substantially less established is whether these components can \textbf{co-evolve over extended time horizons while preserving attribution, independent evaluation, generalization, and reversibility}. A task generator changes the distribution on which the policy is trained; policy updates change the failure modes observed by verifiers; Harness changes alter the evidence available to the model and evaluator; environment changes expose new failure modes; and physical-world failures may in turn induce new tasks, verifiers, data, or model updates. Once several such components are adaptive, their effects are no longer independent.

We therefore formulate self-evolution as a coupled system over task specifications $z$, environments $e$, world and policy programs $W$ and $P$, Harnesses $H$, model state $\theta$, verifiers $\phi$, data $D$, and memory $M$. This formulation differs from component-wise taxonomies in two respects. First, it treats changes to these objects as potentially interdependent rather than as isolated forms of persistent adaptation. Second, it requires evidence that supports attribution across component boundaries. A local improvement is insufficient if it depends on a simultaneously modified evaluator, a narrower task distribution, or an environment-specific shortcut.

This perspective exposes four recurring gaps in prior work. \textbf{First, fixed-component generalization:} improvements are typically measured while most surrounding components remain unchanged, leaving robustness to jointly changing tasks, interfaces, or environments uncertain. \textbf{Second, component-local capability boundaries:} gains in one component may leave system-level bottlenecks unchanged. \textbf{Third, sim-to-real and physical-causality gaps:} improvements in simulation may not transfer reliably, and physical failures are difficult to attribute. \textbf{Fourth, evidence-backed continual evolution:} few systems demonstrate sustained updates with independent validation, regression control, provenance, and rollback. Together, these gaps motivate a coupled and staged approach to self-evolution.

\input{appendix/robocasa_cases}
\input{appendix/phybench_cases}

\end{document}

%% file: tables/hexa_harness.tex
\begin{table}[H]
\centering
\small
\caption{\textbf{Harness comparison on RoboCasa365.} XR-1 runs natively or is called as a tool by a coding agent; both coding agents are driven by GPT-5.6-Sol. Entries are success rates. Each task is evaluated on 50 seeds (18 Atomic-Seen, 16 Composite-Seen, and 16 Composite-Unseen tasks), and Overall pools all trials.}
\label{tab:hexa-harness}
\begin{tabular}{@{}lccc@{}}
\toprule
\textbf{Split} & \textbf{XR-1 (native)} & \textbf{HexaAnything} & \textbf{Codex} \\
\midrule
Atomic-Seen & 78.0\% & 80.9\% & 81.7\% \\
Composite-Seen & 54.8\% & 61.5\% & 59.8\% \\
Composite-Unseen & 34.3\% & 38.3\% & 34.1\% \\
Overall & 56.6\% & \textbf{61.1}\% & 59.5\% \\
\bottomrule
\end{tabular}
\end{table}

%% file: tables/internal_pilot.tex
\begin{table}[H]
\centering
\small
\caption{\textbf{RoboCasa365 Composite-Unseen case study.} Entries are success rates over 100 seeds. HexaAnything, driven by GPT-5.6-Sol, calls the same XR-1 as a tool inside an editable workflow.}
\label{tab:internal-pilot}
\begin{tabular}{@{}lcc@{}}
\toprule
\textbf{Task} & \textbf{XR-1 (native)} & \textbf{HexaAnything} \\
\midrule

LoadKebabSandwich & 17.0\% & 48.0\% \\
PortionHotDogs & 22.0\% & 37.0\% \\
WaffleReheat & 53.0\% & 70.0\% \\
\bottomrule
\end{tabular}
\end{table}

%% file: tables/data_evolution.tex
\begin{table}[H]
\centering
\small
\caption{\textbf{Training data of HexaModel v0.1, trained for one epoch.} The agent traces and the 9.3K RoboCasa365 VQA examples are returned by the Harness.}
\label{tab:data-evolution}
\begin{tabular}{@{}llrr@{}}
\toprule
\textbf{Stream} & \textbf{Content} & \textbf{Records} & \textbf{Tokens} \\
\midrule
Agent traces & subgoals, tool calls, verifier outcomes, recovery decisions & 1,010 & 42.4M \\
VQA with CoT & RoboCasa365 state and task-completion judgments; general embodied VQA & 12,075 & 41.7M \\
Scene captions & RoboCasa365 scene descriptions & 6,764 & 2.0M \\
General data & math, code, and instruction following & 24,491 & 92.5M \\
\midrule
Total & & 44,340 & 178.6M \\
\bottomrule
\end{tabular}
\end{table}

%% file: tables/model_evolution.tex
\begin{table}[H]
\centering
\small
\caption{\textbf{Model evolution on RoboCasa365.} Each row is the planner inside HexaAnything, except XR-1 (native), which runs the VLA without the Harness. Entries are success rates. Each task is evaluated on 50 seeds (18 Atomic-Seen, 16 Composite-Seen, and 16 Composite-Unseen tasks), and Overall pools all trials.}
\label{tab:model-evolution}
\begin{tabular}{@{}lcccc@{}}
\toprule
\textbf{Planner} & \textbf{Atomic-Seen} & \textbf{Composite-Seen} & \textbf{Composite-Unseen} & \textbf{Overall} \\
\midrule
XR-1 (native, no Harness) & 78.0\% & 54.8\% & 34.3\% & 56.6\% \\
\midrule
Qwen3.8-27B (base) & 80.8\% & 61.0\% & 37.3\% & 60.5\% \\
GPT-5.6-Sol & 80.9\% & 61.5\% & 38.3\% & 61.1\% \\
HexaModel v0.1 & 81.0\% & 62.3\% & 39.5\% & 61.7\% \\
\bottomrule
\end{tabular}
\end{table}

%% file: tables/tool_evolution.tex
\begin{table}[H]
\centering
\small
\caption{\textbf{Tool self-evolution on three RoboDojo tasks.} Each cell is the task success rate after the corresponding revision round; the task definition and the evaluator are the same in every round.}
\label{tab:tool-evolution}
\begin{tabular}{@{}lccc@{}}
\toprule
\textbf{Revision round} & \textbf{Fold cloth} & \textbf{Pour vase} & \textbf{Press by number} \\
\midrule
Round 0 & 0.0\% & 40.0\% & 0.0\% \\
Round 1 &  60.0\% & 0.0\% & 0.0\% \\
Round 2 & 80.0\% & 100.0\% & 100.0\% \\
\bottomrule
\end{tabular}
\end{table}

%% file: tables/sci_expriment.tex
\begin{table}[t]
\centering\small
\caption{\textbf{Scientific experiment execution with HexaAnything in simulation.} Entries give mean relative error (\%; lower is better) over valid runs and the valid-run count $n/N$.}
\label{tab:scientific-experiments}
\begin{tabular*}{\linewidth}{@{\extracolsep{\fill}}lccc@{}}
\toprule
\multicolumn{1}{c}{\multirow{2}{*}{Harness + Model}} & \textbf{Hooke's law} & \textbf{Simple pendulum} & \textbf{Coupled oscillators}\\
 & $k$ error $\downarrow$ & $g$ error $\downarrow$ & $\omega_{1,2}$ error $\downarrow$\\
\midrule
HexaAnything + GPT-6-Astra & 2.3\% (10/10) & 1.2\% (10/10) & 0.8\% (10/10)\\
HexaAnything + Opus 5.5 & 1.3\% (10/10) & 1.0\% (10/10) & 1.7\% (10/10)\\
HexaAnything + GPT-5.6-Sol & 4.8\% (9/10) & 1.9\% (8/10) & 1.2\% (10/10)\\
HexaAnything + Qwen3.8-Max & 12.4\% (5/10) & 2.4\% (3/10) & 3.7\% (3/10)\\
\bottomrule
\end{tabular*}
\end{table}

%% file: tables/real_robot_tasks.tex
\begin{table}[H]
\centering
\small
\caption{\textbf{Seven tabletop tasks on the AgileX dual-arm platform,} each run three times with GPT-6-Astra as the execution agent calling URAI tools. Progress is the fraction of the task completed (blocks in the bowl out of six for toss blocks; the operator's completion score for fold clothes). Tokens are the execution agent's output tokens per episode, reasoning included; time is wall-clock from task release to a confirmed outcome and includes model latency, robot motion, and, in tic-tac-toe, the human's moves. Published references use the same model on other hardware (GPT-Policy on its own robot; Robocurve on YAM arms with a 25\% speed cap and at most 20 model calls) and are not same-robot measurements. The first unscrew-bottle-cap trial excludes an initial diagnosis-and-repair phase.}
\label{tab:real-robot-tasks}
\begin{tabular}{@{}lccccl@{}}
\toprule
\textbf{Task} & \textbf{Success} & \textbf{Progress} & \textbf{Output tokens} & \textbf{Time (min)} & \textbf{Published reference (same model)} \\
\midrule
Unscrew bottle cap & 3/3 & 100\% & 1.52k & 5.1 & GPT-Policy R3: 3/3, 17.9 min \\
Tic-tac-toe & 3/3 & 100\% & 3.26k & 5.0 & GPT-Policy R8: 3/3, 13.6 min \\
Block into bowl & 3/3 & 100\% & 443 & 1.0 & Robocurve: 19/20, 2.1k tokens, 2.5 min \\
Hot-dog serving & 3/3 & 100\% & 2.40k & 7.9 & -- \\
Pour blocks & 3/3 & 100\% & 2.29k & 3.1 & -- \\
Toss blocks into bowl & 1/3 & 77.8\% & 4.69k & 8.1 & -- \\
Fold clothes & 1/3 & 83.3\% & 1.41k & 4.1 & -- \\
\bottomrule
\end{tabular}
\end{table}

%% file: tables/related_work.tex
{\scriptsize
\setlength{\tabcolsep}{2.5pt}
\renewcommand{\arraystretch}{1.08}
\begin{longtable}{@{}>{\raggedright\arraybackslash}p{0.18\textwidth}>{\raggedright\arraybackslash}p{0.19\textwidth}>{\raggedright\arraybackslash}p{0.21\textwidth}>{\raggedright\arraybackslash}p{0.17\textwidth}>{\raggedright\arraybackslash}p{0.17\textwidth}@{}}
\caption{\textbf{Mechanism-level comparison of related work.} Rows group representative systems by the artifact whose adaptation is central. ``Fixed boundary'' summarizes the usual reported setup and is not a claim about every implementation.}\label{tab:related-work}\\
\toprule
\textbf{Line of work and representatives} & \textbf{Primary evolving artifact} & \textbf{Persistence and typical fixed boundary} & \textbf{Feedback or evidence} & \textbf{Relation to this work} \\
\midrule
\endfirsthead
\multicolumn{5}{c}{\tablename\ \thetable{} -- continued from the previous page} \\
\toprule
\textbf{Line of work and representatives} & \textbf{Primary evolving artifact} & \textbf{Persistence and typical fixed boundary} & \textbf{Feedback or evidence} & \textbf{Relation to this work} \\
\midrule
\endhead
\midrule
\multicolumn{5}{r}{Continued on the next page} \\
\endfoot
\bottomrule
\endlastfoot
Task, scene, and curriculum generation:\newline
GenSim, RoboGen, SceneSmith, CurricuLLM~\cite{wang2024gensim,wang2023robogen,pfaff2026scenesmith,ryu2025curricullm}
& Task specification $z$, scene/environment $e$, or curriculum
& Generated tasks and scenes persist as training or evaluation assets; the downstream policy, learner, and evaluator are usually external.
& Task solvability, policy performance, or curriculum progress.
& Covers code as world and environment evolution; it does not by itself revise the executing policy and verifier together. \\
\addlinespace
Data and trajectory synthesis:\newline
GenSim2, MimicGen, RoboTwin, HumanoidGen~\cite{hua2024gensim2,mandlekar2023mimicgen,mu2025robotwin,jing2026humanoidgen}
& Data $D$: demonstrations, trajectories, captions, or VQA
& Data is retained for later training; the generator and downstream update rule are usually fixed for a run.
& Dataset scale or quality and downstream policy success.
& Motivates data evolution; our traces link training data to state predicates, recovery, and held-out validation. \\
\addlinespace
Reward and objective search:\newline
Eureka, DrEureka, REvolve~\cite{ma2024eureka,ma2024dreureka,hazra2025revolve}
& Reward code, domain randomization, or preference-derived objective
& Candidate objectives are iterated within training; the policy and evaluation boundary typically remain specified externally.
& Rollout return or success plus automated or human feedback.
& Shows objective evolution, but a training reward is not an independent verifier $\phi$. \\
\addlinespace
Evaluation and verifier generation:\newline
RoboPlayground, AutoEval, SimFoundry~\cite{wang2026roboplayground,zhou2025autoeval,ranawaka2026simfoundry}
& Task specifications, resets, success predicates, and evaluator
& Evaluation artifacts can be reused across policies; policy training is usually held fixed during comparison.
& Terminal success, process evidence, reset reliability, or evaluator agreement.
& Directly informs $\phi$ and provenance; our verifier is coupled to recovery and regression tests while remaining independent. \\
\addlinespace
Programmatic policy synthesis:\newline
Code as Policies, RoboCodeX, RoboScript, RoboCoder, CaP-X~\cite{liang2023code,mu2024robocodex,chen2024roboscript,li2024robocoder,fu2026cap}
& Executable policy programs and skill compositions
& Programs execute per task or are reused as skills; the base model, tool semantics, and evaluator are generally fixed.
& Execution success, task completion, and code or skill validity.
& Closest precedent for code as policy; our world program makes intermediate state and constraints explicit. \\
\addlinespace
Tools, interfaces, and agent runtimes:\newline
ReAct, SWE-agent, OpenHands, Model Hardware Standard~\cite{yao2023react,yang2024swe,wang2025openhands,anthropic2026mhs}
& Tool schema, agent-computer/robot interface, or hardware driver
& The interface structures observations and actions; model weights and task/evaluation protocols generally remain fixed.
& Tool execution, repository tests, or interface-level correctness.
& Supports modular boundaries and physical adapters, but interface standardization alone is not self-evolution. \\
\addlinespace
Persistent memory and skills:\newline
Voyager, MEMENTO, Playful Agentic Robot Learning~\cite{wang2023voyager,sygkounas2026memento,zhang2026playful}
& Skill library, code snippets, or episodic memory $M$
& Artifacts persist across episodes; the foundation model and tool semantics usually stay unchanged.
& Skill reuse, later-task success, or experience accumulation.
& Demonstrates memory-level evolution; our loop also edits world predicates, workflow, and verifiers. \\
\addlinespace
Embodied Harness evolution:\newline
Guava, Harness VLA, Self-Evolving Embodied Agents, Zetta~\cite{liu2026guava,zhang2026harness,wang2026self,ding2026zetta}
& Harness $H$: routing, critics, recovery, memory, and tool orchestration
& Harness revisions can persist across tasks; the low-level action model or base policy is often frozen for attribution.
& Manipulation success, recovery, or closed-loop execution.
& Closest systems-level comparison; our experiments separate Harness gains from model/data evolution and require independent validation. \\
\addlinespace
Model, training, and research-procedure evolution:\newline
ENPIRE, AutoML-Zero, AlphaEvolve, Darwin G\"odel Machine, Autoresearch~\cite{xiao2026enpire,real2020automl,alphaevolve,zhang2026darwin,karpathy2026autoresearch}
& Parameters or adapters $\theta$, algorithms, training scripts, or agent code
& Updates persist as checkpoints or program versions; task, data, evaluator, and runtime boundaries are usually constrained.
& Training or execution score, experiment results, and candidate selection.
& Provides precedents for model and procedure evolution; our claim concerns coupled physical artifacts, not unrestricted self-modification. \\
\addlinespace
Runtime and physical execution substrate:\newline
CompilerGym, KernelBench, KernelFoundry, Evolution Gym~\cite{cummins2022compilergym,ouyang2025kernelbench,wiedemann2026kernelfoundry,bhatia2021evolution}
& Compiler/runtime code, kernels, morphology, or simulator substrate $e$
& Optimized artifacts persist for a workload or platform; workloads, hardware, and evaluator often remain fixed.
& Compilation correctness, runtime, hardware profiling, or task fitness.
& Extends evolution below the policy layer; our formulation treats physical execution and sim-to-real evidence as part of the boundary. \\
\addlinespace
\textbf{This work}: Code as world + code as policy
& $W,P,H,\phi,D,M,e$ in a staged, versioned loop
& Candidate edits persist only after held-out validation; experiments selectively freeze the action model, tasks, or evaluator for attribution.
& State predicates, trajectory evidence, success/recovery, regression, and transfer.
& Jointly exposes state and workflow, and makes the boundary itself an object of controlled evolution. \\
\end{longtable}
}

%% file: appendix/robocasa_cases.tex
\section{RoboCasa365 Case Studies}
\label{app:robocasa-cases}

\begin{figure}[H]
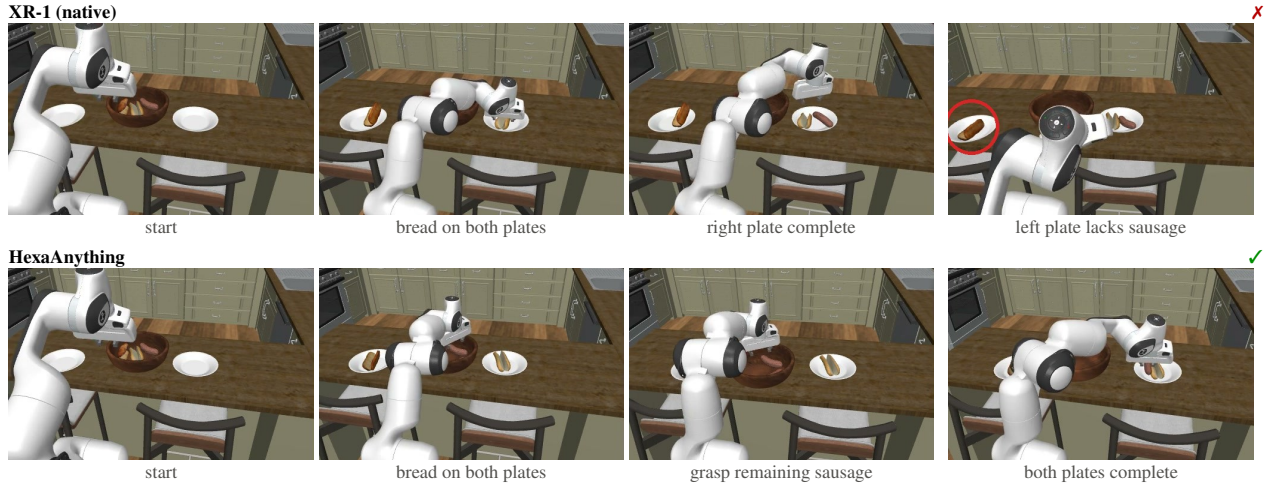

\centering\scriptsize
\setlength{\tabcolsep}{1pt}\renewcommand{\arraystretch}{0.9}
\newcommand{\hd}[3]{\includegraphics[width=0.243\linewidth]{figures/hotdog_case/s#1_#2_#3.jpg}}
\begin{tabular}{@{}cccc@{}}
\multicolumn{4}{@{}p{\linewidth}@{}}{\textbf{XR-1 (native)}\hfill\textcolor{red!70!black}{\ding{55}}} \\
\hd{25}{vla}{0} & \hd{25}{vla}{24} & \hd{25}{vla}{40} & \hd{25}{vla}{108} \\
\color{inkMuted}start & \color{inkMuted}bread on both plates & \color{inkMuted}right plate complete & \color{inkMuted}left plate lacks sausage \\[4pt]
\multicolumn{4}{@{}p{\linewidth}@{}}{\textbf{HexaAnything}\hfill\textcolor{green!55!black}{\ding{51}}} \\
\hd{25}{hexa}{0} & \hd{25}{hexa}{36} & \hd{25}{hexa}{52} & \hd{25}{hexa}{128} \\
\color{inkMuted}start & \color{inkMuted}bread on both plates & \color{inkMuted}grasp remaining sausage & \color{inkMuted}both plates complete \\
\end{tabular}
\caption{\textbf{PortionHotDogs, seed 25.} Native XR-1 completes the right plate and puts bread on the left one, but does not return to the sausage left in the bowl; the left plate ends with bread only (red circle). HexaAnything detects the missing item and issues a grasp for the sausage in the bowl, which XR-1 then places.}
\label{fig:hotdog-25}
\end{figure}

\begin{figure}[H]
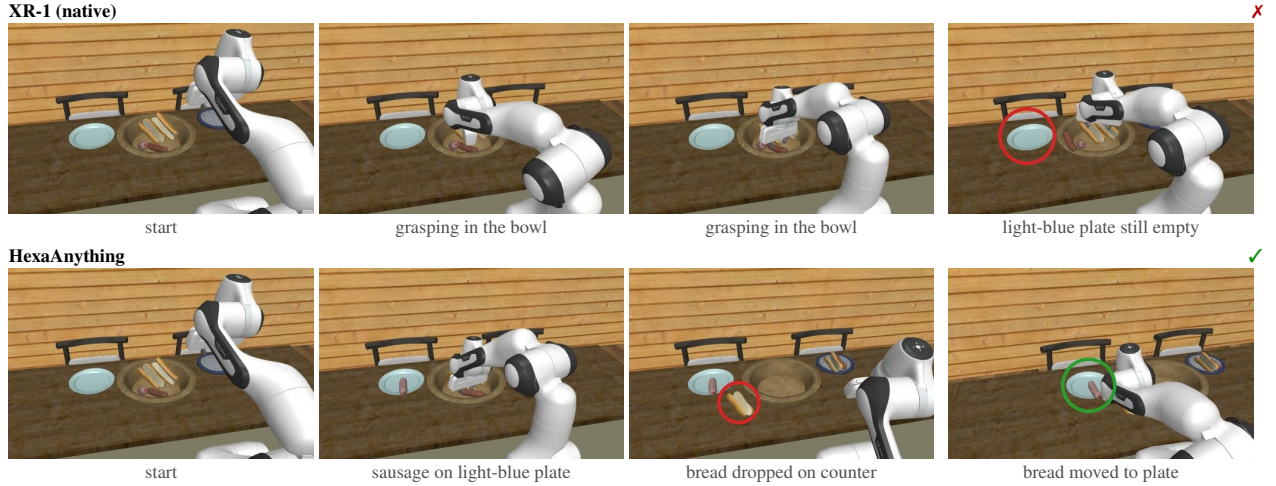

\centering\scriptsize
\setlength{\tabcolsep}{1pt}\renewcommand{\arraystretch}{0.9}
\newcommand{\hd}[3]{\includegraphics[width=0.243\linewidth]{figures/hotdog_case/s#1_#2_#3.jpg}}
\begin{tabular}{@{}cccc@{}}
\multicolumn{4}{@{}p{\linewidth}@{}}{\textbf{XR-1 (native)}\hfill\textcolor{red!70!black}{\ding{55}}} \\
\hd{33}{vla}{0} & \hd{33}{vla}{36} & \hd{33}{vla}{72} & \hd{33}{vla}{108} \\
\color{inkMuted}start & \color{inkMuted}grasping in the bowl & \color{inkMuted}grasping in the bowl & \color{inkMuted}light-blue plate still empty \\[4pt]
\multicolumn{4}{@{}p{\linewidth}@{}}{\textbf{HexaAnything}\hfill\textcolor{green!55!black}{\ding{51}}} \\
\hd{33}{hexa}{0} & \hd{33}{hexa}{68} & \hd{33}{hexa}{104} & \hd{33}{hexa}{128} \\
\color{inkMuted}start & \color{inkMuted}sausage on light-blue plate & \color{inkMuted}bread dropped on counter & \color{inkMuted}bread moved to plate \\
\end{tabular}
\caption{\textbf{PortionHotDogs, seed 33.} Native XR-1 keeps grasping inside the bowl, and the light-blue plate stays empty (red circle). In the HexaAnything run, a bread falls onto the counter beside the bowl (red circle); the agent detects the failed placement and has XR-1 put that bread on the light-blue plate (green circle).}
\label{fig:hotdog-33}
\end{figure}

\begin{figure}[H]
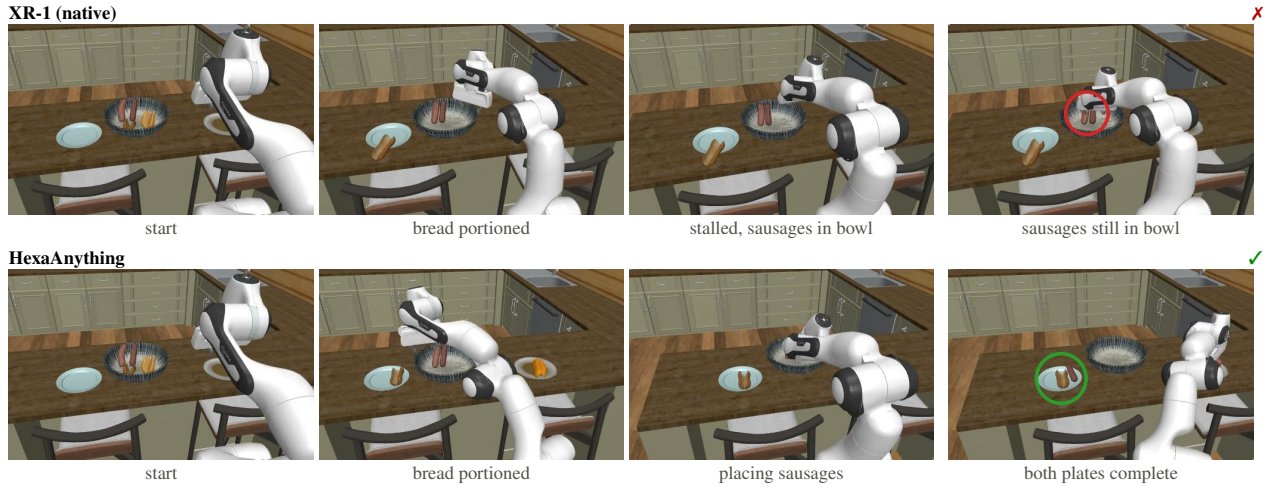

\centering\scriptsize
\setlength{\tabcolsep}{1pt}\renewcommand{\arraystretch}{0.9}
\newcommand{\hd}[3]{\includegraphics[width=0.243\linewidth]{figures/hotdog_case/s#1_#2_#3.jpg}}
\begin{tabular}{@{}cccc@{}}
\multicolumn{4}{@{}p{\linewidth}@{}}{\textbf{XR-1 (native)}\hfill\textcolor{red!70!black}{\ding{55}}} \\
\hd{46}{vla}{0} & \hd{46}{vla}{56} & \hd{46}{vla}{84} & \hd{46}{vla}{108} \\
\color{inkMuted}start & \color{inkMuted}bread portioned & \color{inkMuted}stalled, sausages in bowl & \color{inkMuted}sausages still in bowl \\[4pt]
\multicolumn{4}{@{}p{\linewidth}@{}}{\textbf{HexaAnything}\hfill\textcolor{green!55!black}{\ding{51}}} \\
\hd{46}{hexa}{0} & \hd{46}{hexa}{56} & \hd{46}{hexa}{116} & \hd{46}{hexa}{136} \\
\color{inkMuted}start & \color{inkMuted}bread portioned & \color{inkMuted}placing sausages & \color{inkMuted}both plates complete \\
\end{tabular}
\caption{\textbf{PortionHotDogs, seed 46.} Native XR-1 portions both breads and then stalls with both sausages in the bowl (red circle). HexaAnything checks task progress and issues grasp-and-place commands for the sausages; each plate ends with one bread and one sausage (green circle).}
\label{fig:hotdog-46}
\end{figure}

PortionHotDogs requires moving two breads and two sausages from a bowl so that each of two plates holds one bread and one sausage. On the three episodes below, native XR-1 leaves the task unfinished and HexaAnything completes it from the same initial state (Figures~\ref{fig:hotdog-25}--\ref{fig:hotdog-46}); frames are taken from the recorded episodes.

%% file: appendix/phybench_cases.tex
\section{PhyBench Case Studies}
\label{app:phybench-cases}

This appendix shows example runs of the simple-pendulum and coupled-oscillator tasks in the same form as the Hooke's-law case study of Section~\ref{sec:hooke-case}. Both are taken from the HexaAnything + Opus 5.5 runs summarized in Table~\ref{tab:scientific-experiments} and were selected because they completed the official evaluation with all gates passed and kept complete traces. They illustrate how the agent plans, executes, and analyzes an experiment; they are not intended to represent typical accuracy. Photographs are camera frames from the runs; the masks and orange markers are outputs of the agent's tools, and labels, and circles are added for the figures. Plots are redrawn from the CSV and JSON files the agent wrote during each run.

\subsection{Simple Pendulum}

\begin{figure}[h]
\centering
\includegraphics[width=\linewidth]{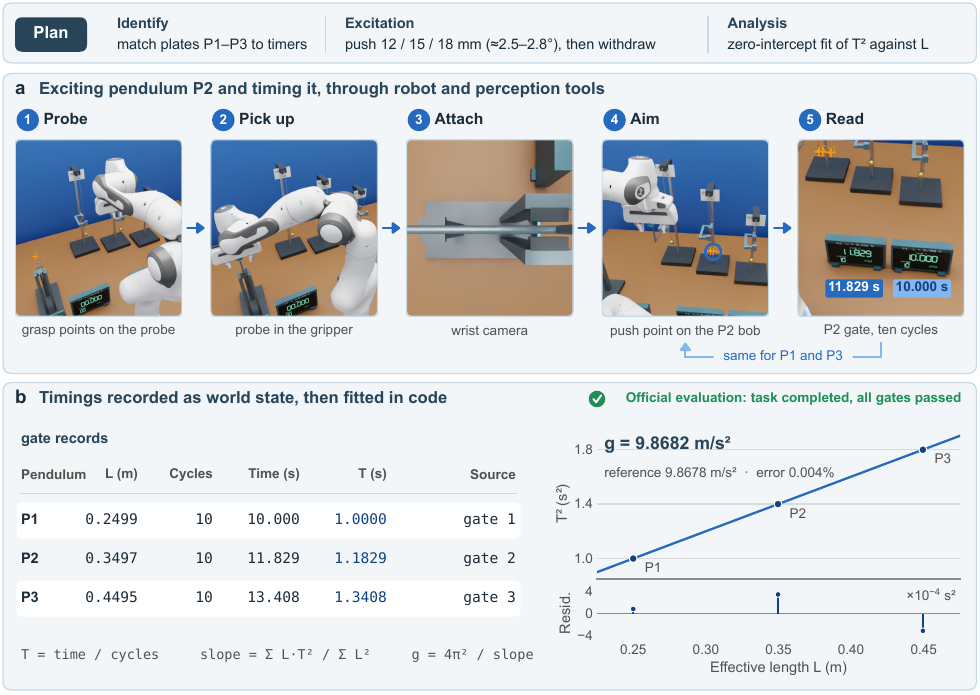}
\caption{\textbf{Simple-pendulum example run.} The plan (top) is written before any action. (a) One excitation, shown for P2: the agent marks grasp points on the magnetic probe and picks it up, the wrist camera shows the attached probe, the agent marks a push point on the bob (circled), pushes and withdraws, and reads the ten-cycle time from the gate timer. The same steps are repeated for P1 and P3. (b) Gate records, each with the gate-timer record it was read from, and the zero-intercept fit of $T^2$ against $L$ over the three pendulums; the lower strip shows residuals.}
\label{fig:case-pendulum}
\end{figure}

\paragraph{Plan.}
The task provides three labeled pendulums with calibrated effective lengths. The agent first matched each label plate to its optical-gate timer. It planned to excite the bobs with the magnetic probe, pushing P1, P2, and P3 by 12, 15, and 18\,mm so that the amplitude stays at about 2.5--2.8$^\circ$, below the $5^\circ$ limit, and to withdraw immediately so that each pendulum swings freely while ten complete cycles are timed. It would then estimate $g$ from $T^2=(4\pi^2/g)L$ over the three lengths.

\paragraph{Execution and analysis.}
After picking up and attaching the probe, the agent marked a push point on each bob, pushed, withdrew, and read the ten-cycle time from the gate timer (Figure~\ref{fig:case-pendulum}a). A zero-intercept fit of $T^2$ against $L$ over the three pendulums gives $g=9.8682$\,m/s$^2$ against a reference of 9.8678\,m/s$^2$, a relative error of 0.004\%, with all residuals below $4\times10^{-4}$\,s$^2$ (Figure~\ref{fig:case-pendulum}b).

\subsection{Coupled Oscillators}

\begin{figure}[h]
\centering
\includegraphics[width=\linewidth]{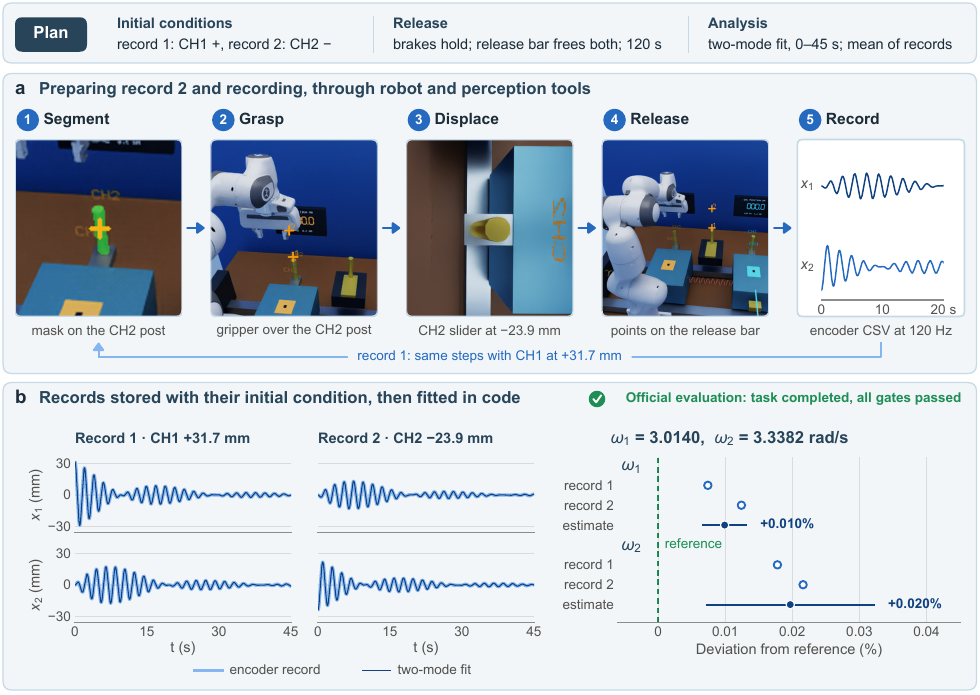}
\caption{\textbf{Coupled-oscillator example run.} The plan (top) is written before any action. (a) Preparation of the second record: the agent segments the CH2 post and selects grasp points on it (gripper above the post), displaces the CH2 slider to $-23.9$\,mm (wrist camera), marks grasp points on the release bar, and pulls it to start a 120\,s two-channel recording. The first record used the same steps with CH1 at $+31.7$\,mm. (b) Both records with the joint two-mode fit over 0--45\,s, and the normal-mode frequencies of each record and of their mean, shown as deviations from the reference.}
\label{fig:case-coupled}
\end{figure}

\begin{table}[H]
\centering\small
\caption{Independent records and estimated normal-mode frequencies in the coupled-oscillator run.}
\label{tab:case-coupled}
\begin{tabular}{lccc}
\toprule
Record & Initial displacement & $\omega_1$ (rad/s) & $\omega_2$ (rad/s) \\
\midrule
1 & CH1 $+31.7$\,mm & 3.0139 & 3.3382 \\
2 & CH2 $-23.9$\,mm & 3.0141 & 3.3383 \\
\midrule
Estimate (mean of records) & & 3.0140 & 3.3382 \\
Reference & & 3.0137 & 3.3376 \\
\bottomrule
\end{tabular}
\end{table}

\paragraph{Plan.}
The task requires two independent initial displacements and free motion during acquisition. The agent designed two records: in the first, only the CH1 slider is displaced in the positive direction while CH2 stays at equilibrium; in the second, only CH2 is displaced in the negative direction. The two initial conditions are linearly independent, and each excites both modes. In each record the sliders are held by their brakes during preparation and released together by pulling the release bar, followed by a 120\,s two-channel recording. Between records, damping is switched on to bring the system to rest.

\paragraph{Execution and analysis.}
For the first record, the agent gripped the CH1 post, released the CH1 brake, translated the slider to $+31.7$\,mm on the encoder, and relocked it. It then armed the recorder and pulled the release bar; Figure~\ref{fig:case-coupled}a shows the same steps for the second record. After exporting the CSV and checking its hash against the instrument, it fitted a joint two-mode model with shared damping (Figure~\ref{fig:case-coupled}b). Before collecting the second record, with CH2 at $-23.9$\,mm, it wrote its analysis rule to its notes: fit the first 45\,s of each record, average the two records, and take the uncertainty as the largest of the between-record spread, the spread across fitting windows, and the formal fit error. The resulting frequencies are 3.0140 and 3.3382\,rad/s against references of 3.0137 and 3.3376\,rad/s, relative errors of 0.010\% and 0.020\% (Table~\ref{tab:case-coupled}).